\documentclass{article}

\usepackage[preprint]{corl_2026} % Use this for the initial submission.

\usepackage[utf8]{inputenc}
\usepackage[T1]{fontenc}
\usepackage{booktabs}
\usepackage{amsmath,amsfonts,amssymb}
\usepackage{nicefrac}
\usepackage{microtype}
\usepackage{graphicx}
\usepackage{multirow}
\usepackage{wrapfig}
\usepackage{float}
\usepackage{makecell}
\usepackage{pifont}
\usepackage{array}
\usepackage{tabularx}
\usepackage{xurl}
\usepackage{bm}
\usepackage{hyperref}
\usepackage{titletoc}
\usepackage{url}
\usepackage{enumitem}

\newcolumntype{Y}{>{\centering\arraybackslash}X}
\newcolumntype{Z}{>{\raggedright\arraybackslash}X}

\definecolor{baseRed}{RGB}{190,70,70}
\definecolor{svaBlue}{RGB}{55,110,180}
\definecolor{lightRed}{RGB}{252,235,235}
\definecolor{lightBlue}{RGB}{232,241,252}
\definecolor{lightGray}{RGB}{245,245,245}
\newcommand{\cmark}{\checkmark}
\newcommand{\xmark}{\ding{55}}

\usepackage[table]{xcolor}
\definecolor{my_green}{RGB}{0,128,0}

\DeclareMathOperator*{\argmax}{argmax}

\title{Look Before You Leap: Distilling Tree Search into Action Evaluation for Frozen VLA Models}

\author{
Xinyi Xie \textsuperscript{1} \quad 
Zican Hu \textsuperscript{1} \quad
Zhanyu Liu \textsuperscript{1} \quad
Yicheng Dong \textsuperscript{1} \quad
Wenhao Wu \textsuperscript{1} \quad \\
\textbf{Zhenhong Sun} \textsuperscript{\textbf{2}} \quad 
\textbf{Haoran Li} \textsuperscript{\textbf{3}} \quad
\textbf{Chunlin Chen} \textsuperscript{\textbf{1}} \quad
\textbf{Zhi Wang} \textsuperscript{\textbf{1}} \quad
\textbf{Pichao Wang} \textsuperscript{\textbf{4}} \quad
\\
\textsuperscript{1} Nanjing University \\
\textsuperscript{2} Australian National University \\
\textsuperscript{3} Institute of Automation, Chinese Academy of Sciences \\
\textsuperscript{4} Nvidia
\\
\texttt{\{xinyixie,zicanhu,zhanyuliu,yichengdong,wenhaowu\}@smail.nju.edu.cn}\\
\texttt{zhenhong.sun@anu.edu.au} \quad
\texttt{lihaoran2015@ia.ac.cn}\\
\texttt{\{clchen,zhiwang\}@nju.edu.cn} \quad
\texttt{pichaowang@gmail.com}
}

\begin{document}
\maketitle
\footnotetext[4]{This work does not relate to the author's position at Nvidia.}

\begin{abstract}
Vision-Language-Action (VLA) models acquire broad embodied capabilities through large-scale pretraining, yet their generalization remains far more fragile than that of LLMs and VLMs. 
The prevailing remedy, post-training via supervised fine-tuning or reinforcement learning, improves task-specific performance but narrows the generalist capability that makes pretraining valuable. 
We identify a key bottleneck: VLA failures stem not only from action \textit{generation} but also from action \textit{evaluation}. 
A diagnostic pass@k study confirms that frozen VLAs already contain competent behaviors in their output distribution, with overall success rates rising from 33\% at pass@1 to 92\% at pass@32.
Inspired by this, we propose \textbf{SVA} (\textbf{S}earch, \textbf{V}alue, and \textbf{A}ct), a simple framework that equips frozen VLA policies with long-term consequence awareness. 
SVA first uses Monte-Carlo tree search in simulation to sample and explore the VLA's output distribution under a finite search budget, collecting diverse trajectories annotated with empirical returns; this knowledge is then distilled into a lightweight Q-value model that predicts the expected consequence of candidate actions; at deployment, the frozen VLA proposes multiple candidates and the evaluator selects the one with the highest uncertainty-regularized Q-value, requiring no simulator access. 
By decoupling action proposal from consequence evaluation, SVA preserves the generalization capacity of the VLA backbone while improving task success. 
Across embodied benchmarks, SVA consistently improves generalization on unseen tasks and exhibits strong test-time scaling; across real-robot manipulation tasks, it improves average success from 42.8\% to 55.6\% (\textbf{+12.8 points}).
Strikingly, SVA enables a 9B VLA to outperform a 27B VLA by \textbf{7 points} at \textbf{27\% lower inference latency}, suggesting that scaling test-time evaluation is more cost-effective than scaling model size.
\end{abstract}

\section{Introduction}\label{sec:intro}
Vision-Language-Action (VLA) models have emerged as a promising paradigm for building generalist robotic agents, leveraging large-scale pretraining on diverse vision-language and robotic data to acquire broad embodied capabilities~\citep{brohan2022rt,zitkovich2023rt}. 
Despite encouraging progress from models such as OpenVLA~\citep{kim2024openvla}, $\pi_{0.5}$~\citep{intelligence2025pi_}, and GR00T~\citep{bjorck2025gr00t}, VLAs remain far more fragile than their LLM/VLM counterparts~\citep{achiam2023gpt}, frequently failing on tasks that lie only modestly outside their training distribution~\citep{li2024simpler}. 
The prevailing remedy is post-training, via supervised fine-tuning (SFT) on curated demonstrations~\citep{team2024octo} or reinforcement learning (RL) with environment rewards~\citep{chen2025conrft,zang2025rlinf}, both of which require updating the VLA backbone. 
Yet updating a multi-billion-parameter backbone is computationally expensive~\citep{wen2025tinyvla,pertsch2025fast}, and the resulting specialist policies tend to narrow the generalist capacity acquired during pretraining~\citep{kim2025fine}.
This tension between task performance and generalization preservation is particularly acute for VLAs, whose generalization is already far more limited than that of LLMs/VLMs and far more costly to lose~\citep{liu2023libero}. 
How to strengthen VLA performance without sacrificing its hard-won generalist capabilities is a pressing challenge.

\begin{wrapfigure}{r}{0.48\textwidth}
  \vspace{-2.5em}
  \centering 
  \includegraphics[width=\linewidth]{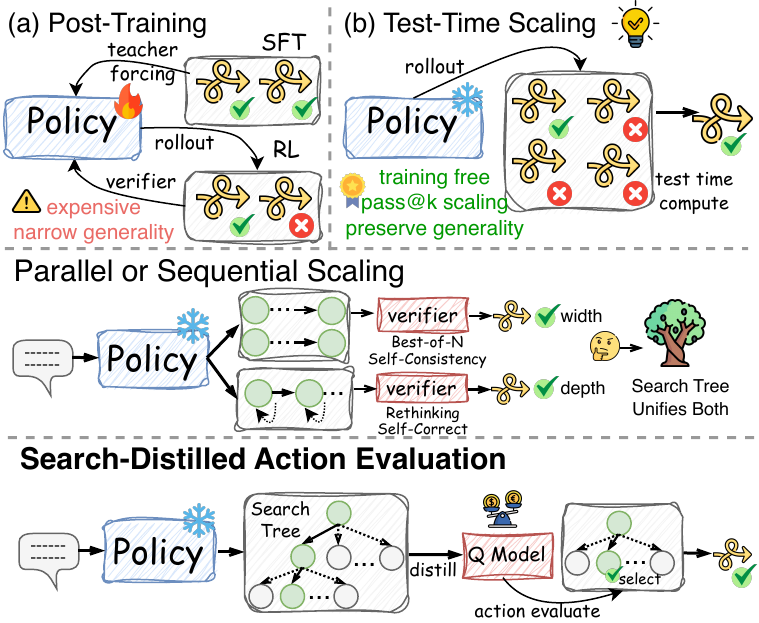}
  % \caption{\textbf{Post-training vs.\ test-time scaling.} Post-training is expensive and narrows generality, whereas test-time scaling is training-free and preserves it. A search tree unifies parallel (width) and sequential (depth) scaling, motivating SVA.}
  \caption{Post-training vs. test-time scaling.}
  \label{fig:concept}
  \vspace{-1em}
\end{wrapfigure}

We argue that the root cause of VLA deployment failures lies not only in action generation but also in the absence of action evaluation, i.e., the inability to anticipate the consequences of a proposed action before execution. 
VLAs are trained to imitate, not to evaluate: they produce locally plausible actions given the current observation~\citep{black2024pi_0}, but receive no signal about what an action leads to (a successful grasp, a collision, or an irrecoverable state)~\citep{de2019causal,codevilla2019exploring}, and no signal about whether a different action might yield a better outcome in the long term~\citep{nakamoto2025steering,sutton1998reinforcement}.
Sec.~\ref{sec:preliminary_passk} provides a diagnostic pass@k study to support this view: when a frozen VLA is allowed multiple independent attempts, success rates rise dramatically from 33\% at pass@1 to 92\% at pass@32.
The analysis reveals that high-quality, task-completing actions already reside within the VLA's output distribution; the model simply cannot distinguish them from mediocre or harmful alternatives it produces alongside. 
\textit{Beyond generation, evaluation is an equally critical yet overlooked bottleneck.}

% A complementary and increasingly compelling perspective is test-time scaling: instead of investing more compute into training larger or more specialized models, one allocates additional computation during inference to improve decision quality. Recent results in large language models (LLMs) and vision-language models (VLMs) domains have demonstrated that scaling test-time compute optimally can yield performance gains rivaling those of model scaling~\citep{chen2021evaluating,snell2025scaling}, including chain-of-thought reasoning~\citep{}, self-consistency reasoning~\citep{}, and tree-of-thought search~\citep{}. This raises a natural yet largely unexplored question: \textit{Can test-time scaling similarly unlock the potential of VLA models for embodied decision-making, offering a training-free route to stronger robotic performance?}

% We first conduct a preliminary study evaluating the Pass@k metric across diverse tasks on multiple VLA benchmarks using popular open-source VLA models. Our results reveal a striking test-time scaling effect: allowing multiple attempts substantially increases the probability of task success, and the scaling gain diminishes with increasing $k$ using naive independent sampling. These results affirm the value of test-time scaling for VLAs and point to a clear opportunity: designing scaling strategies that allocate test computation more intelligently.

This diagnosis motivates a fundamentally different strategy for VLA's policy improvement. Rather than rewriting the VLA's parameters to generate better actions at the cost of generalization, we can equip it with a lightweight consequence evaluator that judges candidate actions by their predicted long-term outcomes. 
The VLA backbone remains entirely frozen, preserving its generalist capacity, while the evaluator provides the missing ``look before you leap'' capability: the ability to foresee which candidate action is most likely to lead to task success.

We draw inspiration from the Bitter Lesson~\citep{sutton2019bitter}: impactful advances in AI stem from methods that scale computation through \textit{search} and \textit{learning}. 
We propose \textbf{SVA} (\textbf{S}earch, \textbf{V}alue, and \textbf{A}ct), a simple three-stage framework that elicits policy improvement for frozen VLAs.
In \textbf{Search}, we employ Monte-Carlo tree search (MCTS) in simulation to sample and explore the VLA's output distribution under a finite search budget, collecting diverse trajectories annotated with empirical returns through look-ahead search.
% The collected data, including both successes and failures, provides rich contrastive supervision for learning action consequences.
In \textbf{Value}, we distill the knowledge obtained from MCTS into a lightweight Q-value model that predicts the expected consequence of executing a candidate action.
Built on a small VLM backbone, this consequence evaluator compresses the expensive search process into a fast-to-evaluate function that generalizes across states and tasks.
In \textbf{Act}, the frozen VLA proposes $N$ candidate actions, and the evaluator selects the one with the highest uncertainty-regularized Q-value.
SVA provides a natural and tunable mechanism for \textit{test-time scaling}: increasing $N$ improves the probability of selecting a high-quality action, with inference latency scaling sub-linearly in $N$ (see Sec.~\ref{sec:candidate-latency}).
% \zicanhu{Beyond simulation, we study evaluator transfer across manipulation tasks on a dual-arm AgileX PiperX, adapting the evaluator with real-robot trajectories while keeping the proposer fixed during evaluation (Sec.~\ref{sec:real-robot}).}
To test whether these gains extend to physical deployment, we evaluate SVA on five manipulation tasks using a dual-arm Agilex PiperX. Under a controlled protocol with the same frozen proposer for both methods, SVA improves mean success from 42.8\% to 55.6\% (\textbf{+12.8 points}; 50 trials per task, with 25 under each scene condition), demonstrating the effectiveness and robustness of evaluator-guided action selection on real hardware (see Sec.~\ref{sec:real-robot}).

% SVA enables real-time action evaluation without simulator access, making it readily deployable in real-world settings, and naturally supports adaptive compute allocation across tasks of varying difficulty.

In summary, our contributions are threefold:
\begin{itemize}[itemsep=0.1em, leftmargin=1.25em]
\vspace{-0.5em}
\item \textbf{Problem.} We identify an evaluation bottleneck in frozen VLA policies: high-quality actions are often present in the output distribution, but single-shot generation cannot reliably pick them out.
\item \textbf{Method.} We propose a search-and-learning framework that distills simulation-based tree search into a value model, enabling real-time action evaluation without simulator access at deployment.
\item \textbf{Results.} SVA delivers consistent gains across embodied benchmarks and real-robot tasks, exhibits strong test-time scaling, and serves as an effective alternative to costly policy fine-tuning; notably, a 9B VLA with SVA outperforms a 27B VLA by 7 points at 27\% lower inference latency. 
\end{itemize}

%=================================================
%=================================================
%=================================================
\section{Related Work}\label{sec:related}
\textbf{VLA Models.}
A growing body of work adapts the success of LLMs/VLMs into embodied domains for generalist robotic control~\citep{li2026simplevla}, with a diverse family of VLAs emerging, including RT-1/RT-2~\citep{brohan2022rt,zitkovich2023rt}, OpenVLA~\citep{kim2024openvla}, $\pi$-series~\citep{black2024pi_0,intelligence2025pi,intelligence2025pi_}, GR00T~\citep{bjorck2025gr00t}, and Octo~\citep{team2024octo}.
Despite this progress, VLAs remain markedly more fragile than their LLM/VLM counterparts, frequently failing on tasks that lie only modestly outside their training distribution~\citep{li2024simpler}.
This fragility motivates a large body of post-training research, mainly SFT on curated data~\citep{pertsch2025fast,wen2025tinyvla} or RL with environment rewards~\citep{xiao2026selfimproving,li2026simplevla,yang2024robot}.
However, these methods share two drawbacks: updating billion-parameter backbones is computationally costly~\citep{kim2025fine,zhao2026simreal}, and specialist fine-tuning often narrows generalist capabilities acquired during pretraining~\citep{liu2023libero,hancock2026actions}.
In contrast, SVA keeps the VLA backbone frozen and steers it via a lightweight external evaluator, avoiding costly backbone updates while preserving its generalist capabilities.

\textbf{Action Evaluation for Robot Policies.}
This complementary line learns a verifier to guide generation, which has proven effective in LLM reasoning, where outcome- and process-reward models~\citep{lightman2024let,uesato2022solving} combined with search~\citep{yao2023tree,wan2024alphazero,zhang2024rest} boost a frozen generator. 
A growing body of work brings this idea to robot policies~\citep{huang2023inner,liang2026adaptive,ayanthi2026rever}.
% PA-RL~\citep{mark2024policy} runs global-local optimization to maximize the action critic
SayCan~\citep{ichter2022do} grounds LLM-proposed plans by scoring candidate skills with an affordance function, V-GPS~\citep{nakamoto2025steering} learns a value function via offline RL pretraining on re-annotated trajectories, RoboMonkey~\citep{kwok2025robomonkey} trains an action preference model using synthetic preference data, and V-VLAPS~\citep{ren2026vvlaps} learns a value model to guide tree search at deployment.
% \citep{wu2025you} uses a pretrained VLM to score the alignment between action and text plan.
Other methods train the action verifier by SFT on synthetic reasoning data in VeGAS~\citep{singhi2026think}, offline RL on a curated dataset in Hume~\citep{song2025hume}, contrastive learning in CoVer-VLA~\citep{kwok2026scaling}, probability matching in TACO~\citep{yang2025steering}, or evolutionary diffusion in VLA-Pilot~\citep{li2026towards}, among others~\citep{mark2024policy,wu2025you}.
Further, MG-Select~\citep{jang2026verifierfree} leverages the model's internal properties to score actions without an external verifier.
World model-based methods~\citep{yang2024learning,zhou2024robodreamer,zhou2025dino,neary2025improving} evaluate actions by rolling them out in a learned world model, but incur high inference cost and accumulated model error over long horizons.
In contrast, SVA complements these approaches along three axes: it learns from MCTS rollouts rather than offline RL or synthetic data, producing consequence-aware signals rather than single-step preference, and obtains long-horizon estimates without costly world-model rollouts.

%=================================================
%=================================================
%=================================================
\section{Diagnosing the VLA Bottleneck: Generation or Evaluation?}
\label{sec:preliminary_passk}
We argue that VLA failures arise not only from action \textit{generation} but also from the absence of action \textit{evaluation}. 
% A direct consequence of this view is that a frozen VLA, even when it fails on single-shot execution, may already contain successful behaviors within its output distribution. 
To this end, we conduct a diagnostic pass@k study to ask how often at least one out of $k$ rollouts succeeds, probing the latent policy capability by isolating evaluation from generation.

\begin{wrapfigure}{r}{0.6\textwidth}
  \vspace{-0.5em}
  \centering 
  \includegraphics[width=\linewidth]{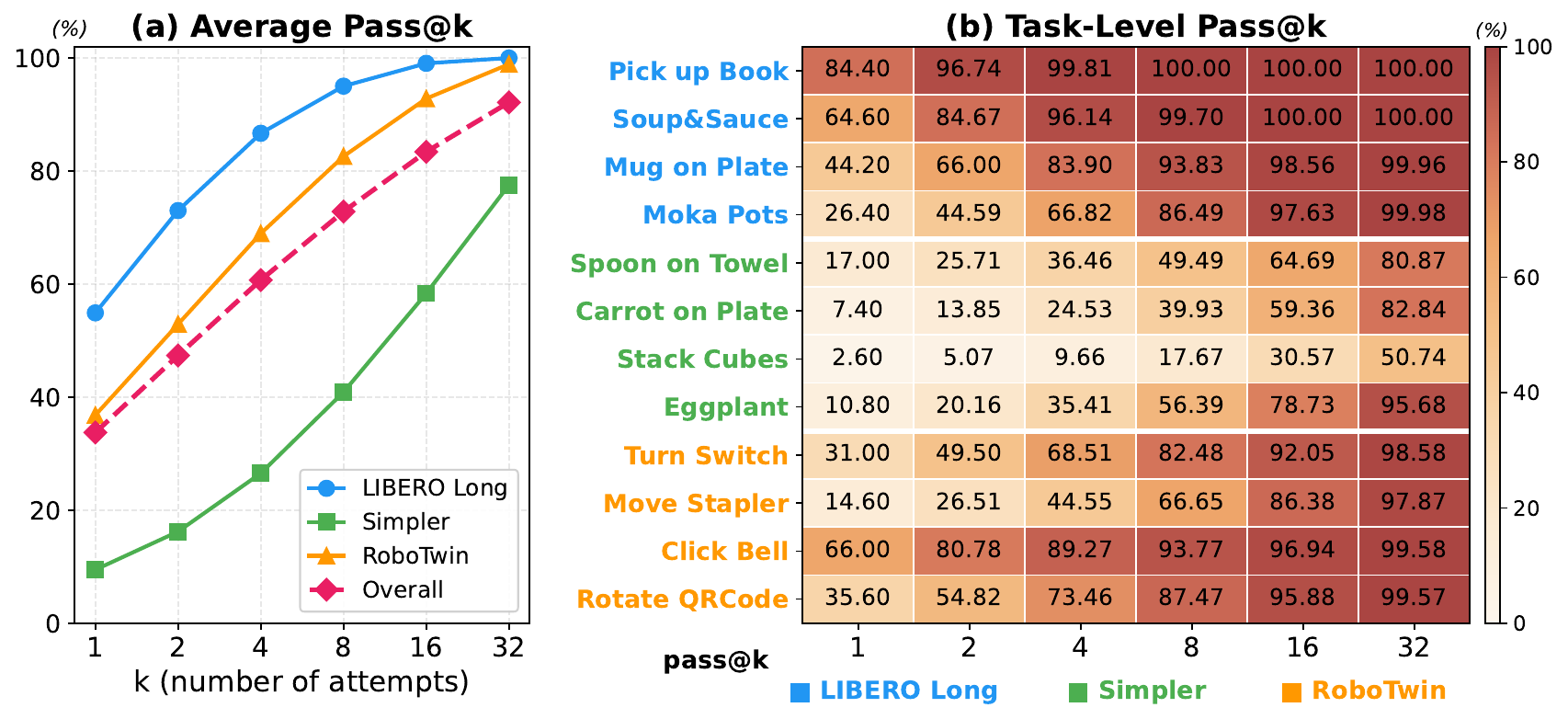}
  \caption{Pass@k results across embodied benchmarks.}
  \label{fig:pre_study}
  \vspace{-0.5em}
\end{wrapfigure}

\textbf{Model, Benchmark, and Evaluation Protocol.}
We evaluate pass@k behavior of OpenVLA on \textsc{Simpler} and \textsc{Libero}, and $\pi_{0.5}$ on \textsc{RoboTwin}.
The three embodied benchmarks cover a range of task structures, including long-horizon object rearrangement, tabletop manipulation, and precise interaction tasks. 
For each task, we sample $N=50$ independent rollouts from the policy distribution with $c$ successes and compute $\mathrm{Pass}@k = \mathbb{E}\bigl[1 - \binom{N-c}{k} / \binom{N}{k}\bigr]$, which estimates the probability that at least one of $k$ independent attempts succeeds~\citep{chen2021evaluating}.
Fig.~\ref{fig:pre_study} shows the diagnostic results.

%================================================
\iffalse 
\begin{table}[t]
\centering
\caption{Benchmark-level average pass@$k$ results. Values are averaged over tasks within each benchmark.}
\label{tab:passk_benchmark_avg}
\resizebox{\linewidth}{!}{
\begin{tabular}{lcccccc}
\toprule
\textbf{Benchmark} & \textbf{pass@1} & \textbf{pass@2} & \textbf{pass@4} & \textbf{pass@8} & \textbf{pass@16} & \textbf{pass@32} \\
\midrule
\textsc{Libero Long} & 0.3594 & 0.5595 & 0.7629 & 0.9137 & 0.9873 & 1.0000 \\
\textsc{Simpler}    & 0.3335 & 0.5313 & 0.6771 & 0.8021 & 0.8750 & 0.9375 \\
\textsc{RoboTwin}   & 0.3475 & 0.5050 & 0.6500 & 0.7550 & 0.8375 & 0.9150 \\
\midrule
\textbf{Overall}    & \textbf{0.3468} & \textbf{0.5319} & \textbf{0.6967} & \textbf{0.8236} & \textbf{0.8999} & \textbf{0.9508} \\
\bottomrule
\end{tabular}
}
\end{table}
\fi 
%================================================

%========================================================
\textbf{Observation 1: Successful Behaviors Already Exist in the Output Distribution of Frozen VLAs.}
The probability of obtaining at least one success increases sharply with the number of attempts, with the same trend holding on every task.
The average success rate rises from 33\% at pass@1 to 92\% at pass@32, an absolute gain of \textbf{58 points}.
Frozen VLAs, despite failing under single-shot execution due to stochastic sampling or compounding execution errors, assign non-trivial probability mass to successful behaviors within their output distribution. 
The deployment bottleneck thus does not lie in the absence of competent actions, but in the inability to identify them before execution. 
This reframes VLA's failure mode as a deficiency in action evaluation rather than solely in action generation.

%========================================================
\textbf{Observation 2: The Evaluation Bottleneck Is Most Severe on Tasks of Intermediate Difficulty.}
Already-easy tasks saturate quickly: \texttt{Pick up Book/Soup and Sauce in Basket} increase from $0.84/0.64$ to $0.99/0.96$ at only pass@4, leaving limited room for further gains.
Tasks that exceed the base policy's capability exhibit limited gains regardless of sampling budget: \texttt{Stack Cubes} only improves from $0.02$ to $0.50$ at pass@$32$.
The practical value of evaluation concentrates in the intermediate-difficulty regime, where the evaluator can efficiently translate latent competence into reliable execution: \texttt{Turn Switch/Rotate QRCode} start from $0.31/0.35$ to near $1$ at pass@32.

%========================================================
\textbf{Observation 3: Diminishing Marginal Returns Suggest the Importance of Smarter Scaling Strategies.}
Although pass@$k$ improves monotonically with $k$, the marginal gain shrinks rapidly as the sampling budget grows: $\Delta_{1\rightarrow2}\!=\!0.13,\Delta_{2\rightarrow4}\!=\!0.13,\Delta_{4\rightarrow8}\!=\!0.12,\Delta_{8\rightarrow16}\!=\!0.10,\Delta_{16\rightarrow32}\!=\!0.08$.
This diminishing-return pattern exposes inefficiency of naive independent sampling: while drawing more rollouts can surface better trajectories, the informational gain per sample shrinks rapidly even as environment-interaction cost grows linearly. 
A practical scaling strategy should therefore allocate additional computation more efficiently, such as evaluating promising actions before execution.

\section{Look Before You Leap: Addressing the Evaluation Bottleneck}
\begin{figure}[tb]
    \centering
    % \framebox[\linewidth]{\rule{0pt}{6cm}\textit{Method overview figure (TODO)}}
    \includegraphics[width=0.9\linewidth]{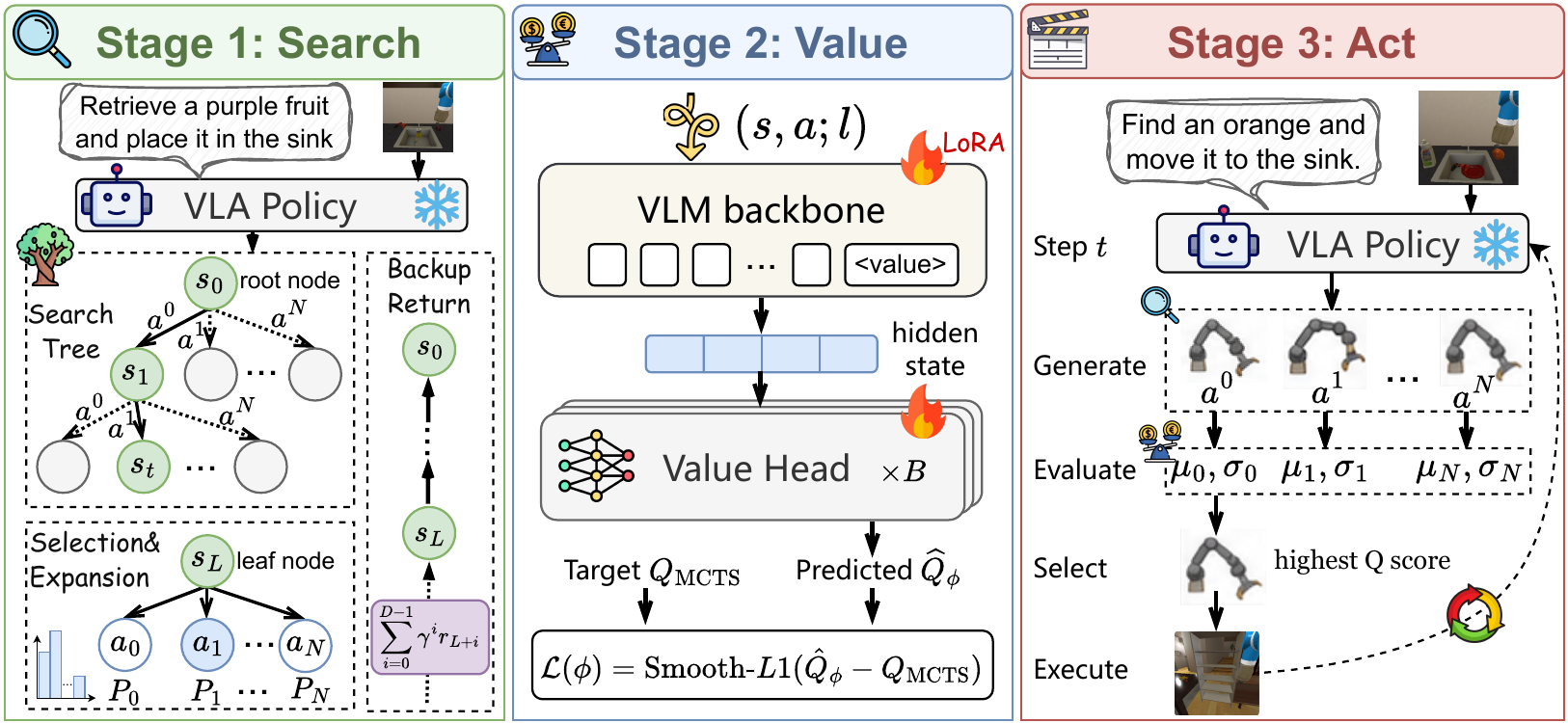}
    % \caption{SVA's overall pipeline.}
    \caption{\textbf{Overview of SVA.} \textbf{(a) Search:} MCTS explores the frozen VLA's policy distribution in simulation, collecting trajectories with empirical returns. \textbf{(b) Value:} A lightweight Q-model is distilled from the searched data to predict action consequences. \textbf{(c) Act:} The frozen VLA proposes $N$ candidates and the Q-model selects the best one without simulator access.}
    \label{fig:method}
\end{figure}

We study language-conditioned embodied tasks in which the agent receives an observation $s_t$ containing images, proprioception, or both, and an instruction $l$. A frozen VLA $\pi_\theta(\cdot\mid s_t;l)$ proposes candidate actions, which may be high-level skills, low-level controls, or action chunks. SVA learns to compare these candidates by their discounted returns, separating action proposal from evaluation. Search collects supervision in simulation, Value fits a deployable evaluator, and Act uses it to select among $N$ candidates without simulator access (Fig.~\ref{fig:method}). Appendix~\ref{app:problem_formulation} gives the full task formulation.

\subsection{Search: Mining Evaluation Signals via MCTS}\label{subsec:mcts}
We use MCTS~\citep{silver2016mastering} to sample and explore the VLA's policy distribution under a finite search budget. Each edge $(s,a)$ stores a return estimate $Q(s,a)$, traversal count $n(s,a)$, return-sample count $m(s,a)$, and prior $P(s,a)$. Selection combines estimated return with prior-weighted exploration. High-level reasoning uses PUCT. Continuous control first visits untried edges in descending prior order, breaking ties randomly, then uses prior-weighted UCT. Updated return estimates guide subsequent traversals.

At an unexpanded node, candidates from Sec.~\ref{subsec:act} become outgoing edges with sequence-frequency or cluster-based priors. After traversing an edge, we evaluate the reached state $s_L$ through frozen policy rollouts, using simulator restoration or reset and replay. Rollouts stop at task termination or a configured cutoff, with zero continuation value. For $D$ executed rollout steps, the return initializes the one-step backward recursion:
\begin{equation}
    G(s_L)=\sum\nolimits_{i=0}^{D-1}\gamma^i r_{L+i},
    \qquad G(s_t)=r_t+\gamma G(s_{t+1}),
    \label{eq:search_returns}
\end{equation}
where $r_t$ is the environment reward and $\gamma$ is the discount factor. Backup proceeds to the root, accumulating rewards and discounts over each action. Each edge averages the return samples at its own starting state. These targets reflect tree-selected actions followed by frozen policy rollouts.

Repeated selection, expansion, rollout, and backup produce supervision from successful and failed searches. Each training sample pairs an observation, instruction, and candidate action with its MCTS-backed return target. These targets remain fixed during evaluator training. Appendix~\ref{app:search_details} provides both selection rules, chunk-level backup, and collection settings.

\subsection{Value: Learning a Deployable Consequence Evaluator}\label{subsec:value}
We distill the search-derived targets into $\hat Q_\phi(s,a;l)$, replacing deployment-time rollouts with direct state-action evaluation. The evaluator uses Qwen3.5-0.8B with LoRA and $B=5$ MLP value heads. High-level skill sequences are serialized as text before a value token. Continuous actions pass through a temporal encoder and are appended with proprioception to the visual--language prefix, followed by a learned Q readout embedding. The contextualized readout feeds the heads, whose mean predicts value and whose standard deviation supplies a disagreement penalty.

Each head regresses to the same target $Q(s,a;l)$. High-level targets are standardized using training-set statistics, while continuous control targets use raw empirical returns. We average the per-head Smooth-$L1$ losses:
\begin{equation}
    \mathcal L_{\mathrm{reg}}(\phi)=\frac{1}{B}\sum\nolimits_{i=1}^{B}
    \operatorname{SmoothL1}\!\left(\epsilon^{(i)}\right),
    \qquad
    \epsilon^{(i)}=\hat Q_\phi^{(i)}(s,a;l)-Q(s,a;l).
    \label{eq:q_regression}
\end{equation}
The objective further averages over samples. Continuous control training can add a pairwise loss over sibling candidates. LoRA, the heads, and added input and readout modules are trained with fixed search targets, without temporal-difference bootstrapping. Appendices~\ref{app:q_architecture} and~\ref{app:value_training} give the architecture, losses, and optimization settings.

\subsection{Act: Evaluation-Guided Action Selection at Test Time}\label{subsec:act}
At each decision, the frozen proposer generates $M$ samples and retains $N\!\leq\! M$ candidates for evaluation. High-level candidates merge identical ordered skill sequences and use their empirical frequencies as priors. Continuous chunks are scored directly with uniform priors or clustered by execution-prefix distance, retaining sampled medoids with smoothed cluster-frequency priors, i.e., $M$ controls sampling and $N$ controls scoring. Appendix~\ref{app:act_details} gives the construction and prior definitions.

The Q-model scores all candidates in a batch. We select the index with the highest regularized score and execute its corresponding action $a_t=a^{(i^*)}$ as 
\begin{equation}
    \begin{gathered}
    i^*=\argmax_{i\in\{1,\ldots,N\}}\Bigl[
    \mu_\phi(s_t,a^{(i)};l)-\lambda_1\sigma_\phi(s_t,a^{(i)};l)
    +\lambda_2\log p(a^{(i)}\mid s_t;l)\Bigr],
    %\\a_t=a^{(i^*)},
    \end{gathered}
    \label{eq:act_selection}
\end{equation}
Here, $\mu_\phi$ and $\sigma_\phi$ are the ensemble mean and population standard deviation, and $\lambda_1,\lambda_2$ control disagreement and prior regularization. The controller executes a configured prefix, stopping early on termination or invalid feedback, then observes again and replans. Prediction length, Q encoding horizon, and executed prefix length are distinct.

The sampling and scoring budgets provide test-time compute knobs while the proposer remains frozen. Shared observation prefixes and batched candidate processing reduce repeated computation in $\pi_{0.5}$ generation and continuous Q scoring. We evaluate the resulting success--latency trade-off and comparison with model-size scaling in Sec.~\ref{sec:candidate-latency}. Appendix~\ref{app:pi05-act-latency} details the accelerated $\pi_{0.5}$ pipeline and latency measurements, with component-wise curves in Fig.~\ref{fig:pi05-act-latency}.

\section{Experiments}
\textbf{Benchmarks.} We evaluate SVA in three settings: 1) \textbf{Embodied Reasoning}, with high-level atomic skill sequences on \textbf{EmbodiedBench}~\citep{yang2025embodiedbench}, using \textsc{EB-Habitat} and \textsc{EB-Navigation}; 2) \textbf{Simulated Manipulation}, with continuous control on \textbf{RoboTwin 2.0}~\citep{chen2025robotwin2}, covering 12 tasks that span short- and long-horizon behavior and both single-arm and bimanual coordination, and on \textbf{SimplerEnv}~\citep{li2024simpler} with the WidowX arm; and 3) \textbf{real robot}, with a dual-arm Agilex PiperX (Sec.~\ref{sec:real-robot}). 
% We report task success rates and equally weighted averages. 
Protocol details are in Appendices~\ref{app:benchmark} and~\ref{app:sva_exp}.
% \textbf{Benchmarks and Metrics.} Our main experiments cover two embodied policy interfaces: high-level reasoning with atomic skill sequences on \textbf{EmbodiedBench}~\citep{yang2025embodiedbench}, and low-level continuous control on \textbf{RoboTwin 2.0}~\citep{chen2025robotwin2}. The high-level evaluation uses \textsc{EB-Habitat} and \textsc{EB-Navigation}; the continuous control evaluation uses 12 manipulation tasks spanning short- and long-horizon behavior, single-arm operation, and bimanual coordination. We report task success rates and equally weighted averages over the specified tasks or benchmark categories. Benchmark and protocol details appear in Appendices~\ref{app:benchmark} and~\ref{app:sva_exp}. Real-world experiments use a dual-arm Agilex PiperX and are reported in Sec.~\ref{sec:real-robot}.

\textbf{Embodied Reasoning Setup} We evaluate GPT-4o, Qwen3.5-4B/9B/27B, and Gemma-4-E4B-it, each with and without SVA. For each EmbodiedBench suite, one Q-model is trained on search data collected with a Qwen3.5 proposer and reused unchanged across all five policies. The evaluator scores ordered atomic skill sequences, with their empirical priors estimated from the frequency of exactly matching sampled sequences.

\textbf{Simulated Manipulation Setup.} On \textbf{RoboTwin 2.0}, we evaluate two frozen proposers, $\pi_{0.5}$~\citep{intelligence2025pi_} and FastWAM~\citep{yuan2026fastwam}, each run directly and with SVA. FastWAM generates action chunks at inference without predicting future video. For $\pi_{0.5}$, we also compare two selectors, V-GPS~\citep{nakamoto2025steering} and TACO~\citep{yang2025steering}, which rank candidates by an offline-RL value and by learned pseudo-counts, respectively. The proposer checkpoint is held fixed when a selector is added, and all six methods share the same 12 tasks. On \textbf{SimplerEnv}, we compare $\pi_0$, $\pi_0$+RoboMonkey~\citep{kwok2025robomonkey}, and $\pi_0$+SVA, with OpenVLA~\citep{kim2024openvla} as a reference generalist. Details are in Appendix~\ref{app:baselines}.

\subsection{Simulation Results}
\label{sec:main-results}

% \subsubsection{EmbodiedBench}\label{sec:embodiedbench-results}
\paragraph{Embodied Reasoning.}
Table~\ref{tab:embodiedbench-main} summarizes results on EB-Habitat and EB-Navigation.
SVA consistently improves success rates across all five backbones, yielding average gains of \textbf{+15.4} on EB-Habitat and \textbf{+13.2} points on EB-Navigation.
% This confirms that SVA acts as a model-agnostic decision-time enhancement rather than being tied to a particular backbone architecture.
These improvements show that the benchmark-specific evaluator can improve average success when paired with the tested proposer families and model sizes.
The most pronounced gains emerge on categories requiring long-horizon planning or visual grounding: Long Horizon (\textbf{+23.34} on Qwen3.5-27B) and Common Sense (\textbf{+28.33} on Qwen3.5-27B), where the base policy tends to commit prematurely to locally plausible but globally suboptimal actions.
By scoring candidate actions with predicted long-term consequences, the Q-model effectively corrects these myopic preferences.
Some categories regress, including GPT-4o on Complex Instruction and Qwen3.5-27B on Spatial Relationship, even though average performance improves.

\begin{table}[t]
\centering
\caption{Success rates (\%) on EmbodiedBench. \textcolor{my_green}{Green}/\textcolor{red}{Red} $\Delta$ denotes gain/drop over the base policy.}
\label{tab:embodiedbench-main}
\setlength{\tabcolsep}{2.5pt}
\renewcommand{\arraystretch}{1.1}
\resizebox{\textwidth}{!}{%
\begin{tabular}{llc cccccc >{\columncolor{cyan!10}}c c ccccc >{\columncolor{cyan!10}}c}
\toprule
\multirow{2}{*}{\textbf{Model}} & \multirow{2}{*}{\textbf{Method}} & \phantom{}
 & \multicolumn{7}{c}{\textbf{EB-Habitat}}
 & \phantom{}
 & \multicolumn{6}{c}{\textbf{EB-Navigation}} \\
\cmidrule(lr){4-10} \cmidrule(lr){12-17}
 & & & \textbf{Base} & \makecell{\textbf{Common}\\\textbf{Sense}} & \makecell{\textbf{Complex}\\\textbf{Instr.}} & \makecell{\textbf{Spatial}\\\textbf{Rel.}} & \makecell{\textbf{Visual}\\\textbf{App.}} & \makecell{\textbf{Long}\\\textbf{Horizon}} & \textbf{Avg.}
 & & \textbf{Base} & \makecell{\textbf{Common}\\\textbf{Sense}} & \makecell{\textbf{Complex}\\\textbf{Instr.}} & \makecell{\textbf{Visual}\\\textbf{App.}} & \makecell{\textbf{Long}\\\textbf{Horizon}} & \textbf{Avg.} \\
\midrule
\multirow{3}{*}{GPT-4o}
  & Base & & 73.33 & 20.00 & 50.00 & 51.67 & 40.00 & 28.33 & 43.89
         & & 47.22 & 30.56 & 51.39 & 38.89 & 44.44 & 42.50 \\
  & SVA  & & \textbf{88.33} & \textbf{36.67} & 48.33 & \textbf{56.67} & \textbf{58.33} & \textbf{46.67} & \textbf{55.83}
         & & \textbf{51.39} & \textbf{50.00} & 45.83 & \textbf{41.67} & \textbf{50.00} & \textbf{47.78} \\
  & $\Delta$ & & \textcolor{my_green}{+15.00} & \textcolor{my_green}{+16.67} & \textcolor{red}{-1.67} & \textcolor{my_green}{+5.00} & \textcolor{my_green}{+18.33} & \textcolor{my_green}{+18.34} & \textcolor{my_green}{+11.94}
         & & \textcolor{my_green}{+4.17} & \textcolor{my_green}{+19.44} & \textcolor{red}{-5.56} & \textcolor{my_green}{+2.78} & \textcolor{my_green}{+5.56} & \textcolor{my_green}{+5.28} \\
\midrule
\multirow{3}{*}{Qwen3.5-4B}
  & Base & & 56.67 & 3.33  & 33.33 & 48.33 & 28.33 & 16.67 & 31.11
         & & 45.83 & 38.89 & 31.94 & 31.94 & 25.00 & 34.72 \\
  & SVA  & & \textbf{96.67} & \textbf{25.00} & \textbf{50.00} & \textbf{56.67} & \textbf{50.00} & \textbf{40.00} & \textbf{53.06}
         & & \textbf{54.17} & \textbf{41.67} & \textbf{45.83} & \textbf{41.67} & \textbf{50.00} & \textbf{46.67} \\
  & $\Delta$ & & \textcolor{my_green}{+40.00} & \textcolor{my_green}{+21.67} & \textcolor{my_green}{+16.67} & \textcolor{my_green}{+8.34} & \textcolor{my_green}{+21.67} & \textcolor{my_green}{+23.33} & \textcolor{my_green}{+21.95}
         & & \textcolor{my_green}{+8.34} & \textcolor{my_green}{+2.78} & \textcolor{my_green}{+13.89} & \textcolor{my_green}{+9.73} & \textcolor{my_green}{+25.00} & \textcolor{my_green}{+11.95} \\
\midrule
\multirow{3}{*}{Qwen3.5-9B}
  & Base & & 71.67 & 10.00 & 48.33 & 55.00 & 35.00 & 36.67 & 42.78
         & & 40.28 & 37.50 & 40.28 & 30.56 & 47.22 & 39.17 \\
  & SVA  & & \textbf{96.67} & \textbf{36.67} & \textbf{63.33} & \textbf{56.67} & \textbf{43.33} & \textbf{46.67} & \textbf{57.22}
         & & \textbf{62.50} & \textbf{59.72} & \textbf{52.78} & \textbf{47.22} & \textbf{58.33} & \textbf{56.11} \\
  & $\Delta$ & & \textcolor{my_green}{+25.00} & \textcolor{my_green}{+26.67} & \textcolor{my_green}{+15.00} & \textcolor{my_green}{+1.67} & \textcolor{my_green}{+8.33} & \textcolor{my_green}{+10.00} & \textcolor{my_green}{+14.44}
         & & \textcolor{my_green}{+22.22} & \textcolor{my_green}{+22.22} & \textcolor{my_green}{+12.50} & \textcolor{my_green}{+16.66} & \textcolor{my_green}{+11.11} & \textcolor{my_green}{+16.94} \\
\midrule
\multirow{3}{*}{Qwen3.5-27B}
  & Base & & 83.33 & 11.67 & 50.00 & 55.00 & 45.00 & 38.33 & 47.22
         & & 40.28 & 51.39 & 44.44 & 37.50 & 16.67 & 38.06 \\
  & SVA  & & \textbf{93.33} & \textbf{40.00} & \textbf{70.00} & 53.33 & \textbf{66.67} & \textbf{61.67} & \textbf{64.17}
         & & \textbf{61.11} & \textbf{56.94} & \textbf{58.33} & \textbf{58.33} & \textbf{52.78} & \textbf{57.50} \\
  & $\Delta$ & & \textcolor{my_green}{+10.00} & \textcolor{my_green}{+28.33} & \textcolor{my_green}{+20.00} & \textcolor{red}{-1.67} & \textcolor{my_green}{+21.67} & \textcolor{my_green}{+23.34} & \textcolor{my_green}{+16.95}
         & & \textcolor{my_green}{+20.83} & \textcolor{my_green}{+5.55} & \textcolor{my_green}{+13.89} & \textcolor{my_green}{+20.83} & \textcolor{my_green}{+36.11} & \textcolor{my_green}{+19.44} \\
\midrule
\multirow{3}{*}{\makecell[l]{Gemma-4-E4B-it}}
  & Base & & 63.33 & 1.67  & 16.67 & 35.00 & 23.33 & 11.67 & 25.28
         & & 37.50 & 25.00 & 38.89 & 30.56 & 11.11 & 28.61 \\
  & SVA  & & \textbf{81.67} & \textbf{11.67} & \textbf{36.67} & \textbf{45.00} & \textbf{31.67} & \textbf{16.67} & \textbf{37.22}
         & & \textbf{41.67} & \textbf{37.50} & \textbf{50.00} & \textbf{38.89} & \textbf{37.50} & \textbf{41.11} \\
  & $\Delta$ & & \textcolor{my_green}{+18.34} & \textcolor{my_green}{+10.00} & \textcolor{my_green}{+20.00} & \textcolor{my_green}{+10.00} & \textcolor{my_green}{+8.34} & \textcolor{my_green}{+5.00} & \textcolor{my_green}{+11.94}
         & & \textcolor{my_green}{+4.17} & \textcolor{my_green}{+12.50} & \textcolor{my_green}{+11.11} & \textcolor{my_green}{+8.33} & \textcolor{my_green}{+26.39} & \textcolor{my_green}{+12.50} \\
\bottomrule  
\end{tabular}
}
\end{table}

\paragraph{Simulated Manipulation.}
For RoboTwin 2.0, Table~\ref{tab:robotwin-main} shows that SVA consistently improves both frozen proposers. For $\pi_{0.5}$, SVA raises the average success rate from 62.72 to \textbf{67.78}, a gain of \textbf{5.06} points, and outperforms V-GPS and TACO by \textbf{3.06} and \textbf{2.45} points, respectively. The improvement appears on 11 of the 12 tasks, with particularly large gains on Click Bell, Place Fan, and Turn Switch. SVA also improves FastWAM from 77.72 to \textbf{79.72} (\textbf{+2.00} points), with gains on 8 of 12 tasks. 

\begin{table}[t]
\centering
\caption{Success rates (\%) on RoboTwin 2.0. Base denotes direct execution of the frozen proposer.}
\label{tab:robotwin-main}
\setlength{\tabcolsep}{2.5pt}
\renewcommand{\arraystretch}{1.1}
\resizebox{\textwidth}{!}{%
\begin{tabular}{ll cccccccccccc >{\columncolor{cyan!10}}c}
\toprule
\textbf{Model} & \textbf{Method} & \makecell{\textbf{Click}\\\textbf{Bell}} & \makecell{\textbf{Move Pill-}\\\textbf{bottle Pad}} & \makecell{\textbf{Move Card}\\\textbf{Away}} & \makecell{\textbf{Pick Dual}\\\textbf{Bottles}} & \makecell{\textbf{Place Bread}\\\textbf{Skillet}} & \makecell{\textbf{Place}\\\textbf{Fan}} & \makecell{\textbf{Place}\\\textbf{Mouse Pad}} & \makecell{\textbf{Press}\\\textbf{Stapler}} & \makecell{\textbf{Scan}\\\textbf{Object}} & \makecell{\textbf{Stack}\\\textbf{Blocks}} & \makecell{\textbf{Stamp}\\\textbf{Seal}} & \makecell{\textbf{Turn}\\\textbf{Switch}} & \textbf{Avg.} \\
\midrule
\multirow{4}{*}{$\pi_{0.5}$} & Base & 20.00 & 69.33 & 81.33 & 83.33 & 70.67 & 67.33 & 40.00 & 76.00 & 50.00 & 96.00 & 50.67 & 48.00 & 62.72 \\
 & +V-GPS & 28.00 & 65.33 & 82.00 & 83.33 & 74.67 & 70.00 & 41.33 & 80.00 & 50.67 & 96.00 & 54.67 & 50.67 & 64.72 \\
 & +TACO & 20.00 & 69.33 & 82.67 & 83.33 & 78.00 & 76.00 & 41.33 & 77.33 & 54.00 & 98.67 & 54.67 & 48.67 & 65.33 \\
 & +SVA & 31.33 & 68.67 & 82.67 & 86.67 & 78.00 & 76.00 & 42.00 & 82.67 & 54.00 & 98.67 & 56.67 & 56.00 & \textbf{67.78} \\
\midrule
\multirow{2}{*}{FastWAM} & Base & 96.67 & 85.33 & 94.00 & 78.00 & 81.33 & 66.67 & 52.00 & 97.33 & 80.00 & 97.33 & 55.33 & 48.67 & 77.72 \\
 & +SVA & 98.67 & 85.33 & 94.67 & 78.00 & 85.33 & 70.67 & 54.00 & 97.33 & 82.67 & 98.67 & 62.67 & 48.67 & \textbf{79.72} \\
\bottomrule
\end{tabular}
}
\end{table}
% \begin{table}[t]
% \centering
% \caption{Success rates (\%) on RoboTwin 2.0. Base denotes direct execution of the frozen proposer.}
% \label{tab:robotwin-main}
% \small
% \setlength{\tabcolsep}{4pt}
% \renewcommand{\arraystretch}{1.1}
% \begin{tabular}{@{}lcccccc@{}}
% \toprule
% & \multicolumn{4}{c}{$\pi_{0.5}$} & \multicolumn{2}{c}{FastWAM} \\
% \cmidrule(lr){2-5}\cmidrule(l){6-7}
% Task & Base & +V-GPS & +TACO & +SVA & Base & +SVA \\
% \midrule
% \texttt{click\_bell}            & 22.00 & 31.33 & 22.00 & 34.00  & 96.00 & 98.00 \\
% \texttt{move\_pillbottle\_pad}  & 70.67 & 66.67 & 70.67 & 70.00  & 85.33 & 85.33 \\
% \texttt{move\_playingcard\_away}& 80.67 & 82.00 & 82.00 & 82.00  & 96.67 & 96.67 \\
% \texttt{pick\_dual\_bottles}    & 82.00 & 82.00 & 82.00 & 84.00  & 78.67 & 78.67 \\
% \texttt{place\_bread\_skillet}  & 66.67 & 68.00 & 70.67 & 71.33  & 84.00 & 86.67 \\
% \texttt{place\_fan}            & 68.67 & 73.33 & 72.67 & 74.67  & 67.33 & 70.67 \\
% \texttt{place\_mouse\_pad}     & 35.33 & 35.33 & 36.67 & 36.67  & 53.33 & 59.33 \\
% \texttt{press\_stapler}        & 74.67 & 78.00 & 79.33 & 80.00  & 91.33 & 91.33 \\
% \texttt{scan\_object}          & 47.33 & 50.00 & 50.67 & 50.67  & 82.00 & 82.67 \\
% \texttt{stack\_blocks\_two}    & 97.33 & 97.33 & 99.33 & 100.00 & 98.00 & 98.67 \\
% \texttt{stamp\_seal}           & 52.67 & 53.33 & 54.00 & 56.67  & 48.67 & 56.67 \\
% \texttt{turn\_switch}          & 54.67 & 59.33 & 56.67 & 61.33  & 41.33 & 42.00 \\
% \midrule
% Avg.                          & 62.72 & 64.72 & 64.72 & 66.78  & 76.89 & 78.89 \\
% \bottomrule
% \end{tabular}
% \end{table}

For SimplerEnv, Table~\ref{tab:simplerenv-main} compares OpenVLA, $\pi_0$, $\pi_0$+RoboMonkey, and $\pi_0$+SVA on four WidowX tasks. SVA raises the $\pi_0$ average from 38.5 to \textbf{50.7}, \textbf{4.4} points above RoboMonkey (46.3). This advantage is consistent with the ranking rule: SVA scores candidates by predicted long-horizon return, whereas RoboMonkey ranks them by local action preference.
\begin{wraptable}[9]{r}{0.5\textwidth}
    \vspace{1.0em}
    \centering
    \caption{Results on SimplerEnv with the WidowX. }
    \label{tab:simplerenv-main}
    \footnotesize
    \setlength{\tabcolsep}{1.5pt}
    \renewcommand{\arraystretch}{1.05}
    \begin{tabular}{@{}lcccc@{}}
    \toprule
    Task & OpenVLA & $\pi_0$ & +RoboMonkey & +SVA \\
    \midrule
    Spoon on Towel & 29.6 & 23.6 & \textbf{29.6} & 27.8 \\
    Carrot on Plate & 3.7 & 23.6 & \textbf{33.3} & 27.8 \\
    Stack Cubes & 0.0 & 37.5 & 40.7 & \textbf{63.9} \\
    Eggplant in Basket & 48.2 & 69.4 & 81.5 & \textbf{83.3} \\
    \midrule
    Average & 20.4 & 38.5 & 46.3 & \textbf{50.7} \\
    \bottomrule
    \end{tabular}
    % \vspace{-1.0em}
\end{wraptable}

\subsection{Real-World Experiments}
\label{sec:real-robot}

\begin{figure}[tb]
  \centering
  \footnotesize
  \vspace{-0.5em}
  % Native LaTeX crops preserve the supplied photographs. Trim order: left, bottom, right, top.
  % Scene photographs are cropped to 2:1 so that the row fills the text width at a low height.
  \begin{tabular*}{\textwidth}{@{\extracolsep{\fill}}ccc@{}}
      \includegraphics[height=0.19\textwidth,trim={50bp 250bp 50bp 150bp},clip]{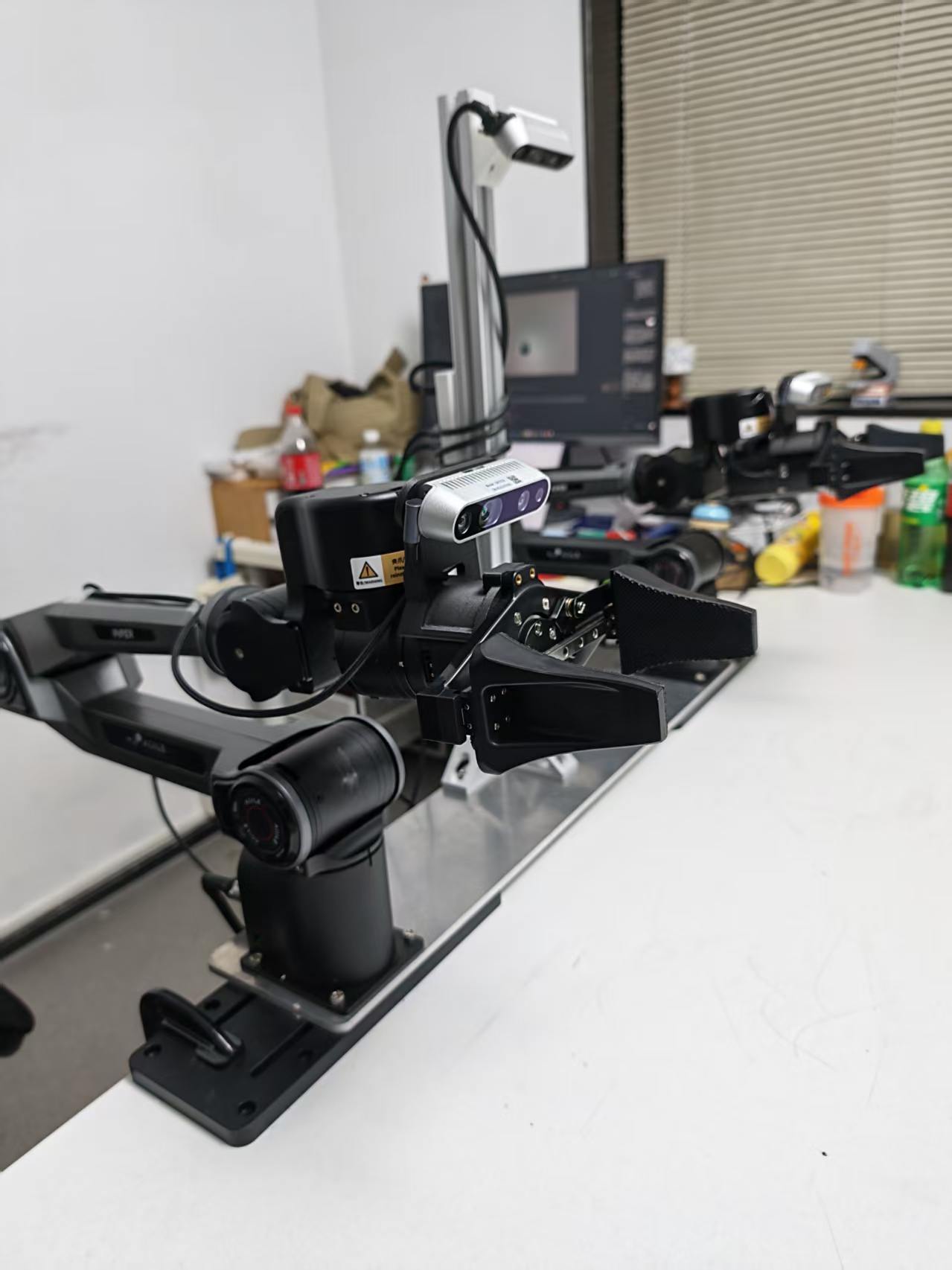}
      & \includegraphics[height=0.19\textwidth,trim={50bp 400bp 50bp 75bp},clip]{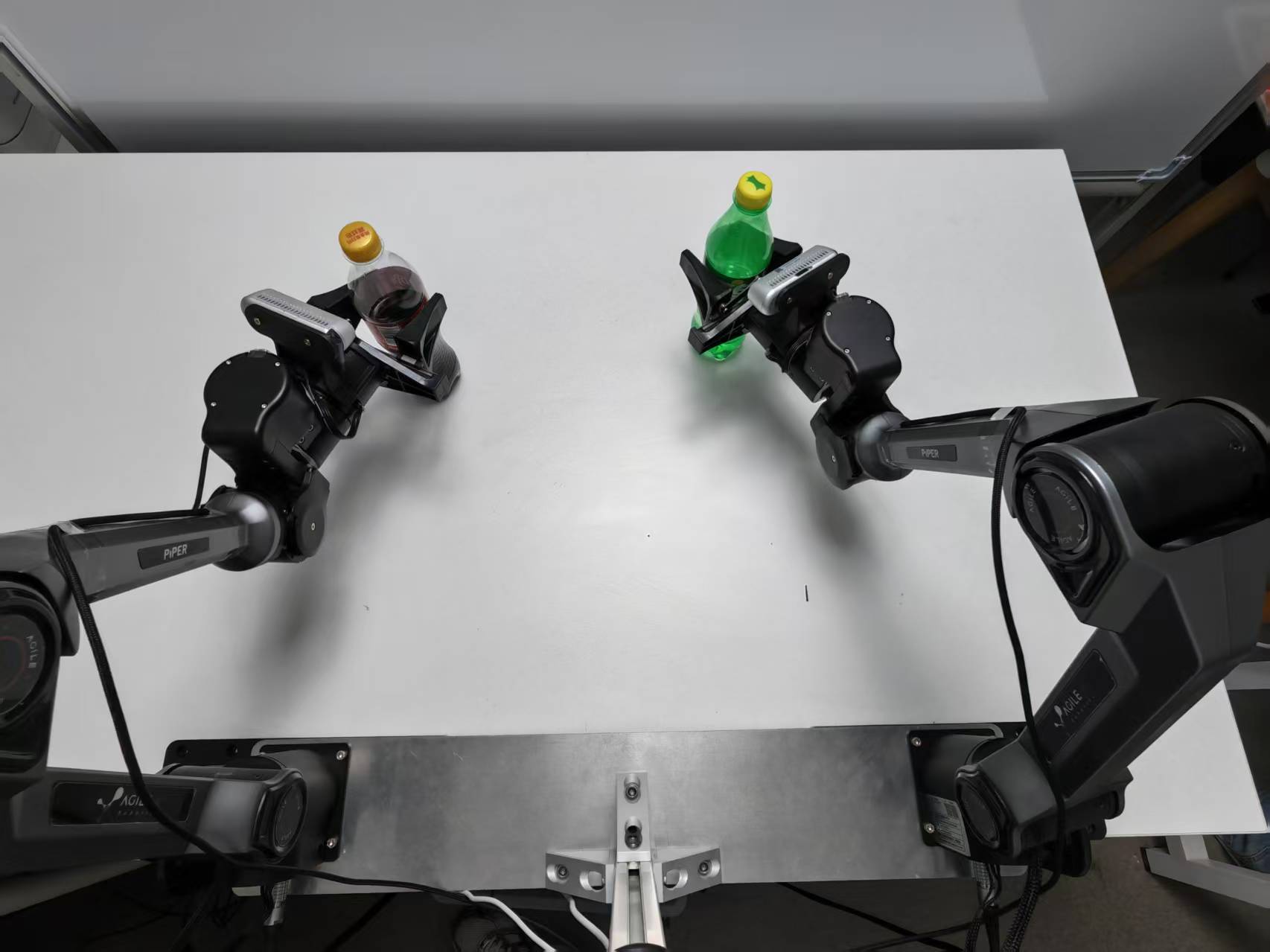}
      & \includegraphics[height=0.19\textwidth,trim={50bp 400bp 50bp 75bp},clip]{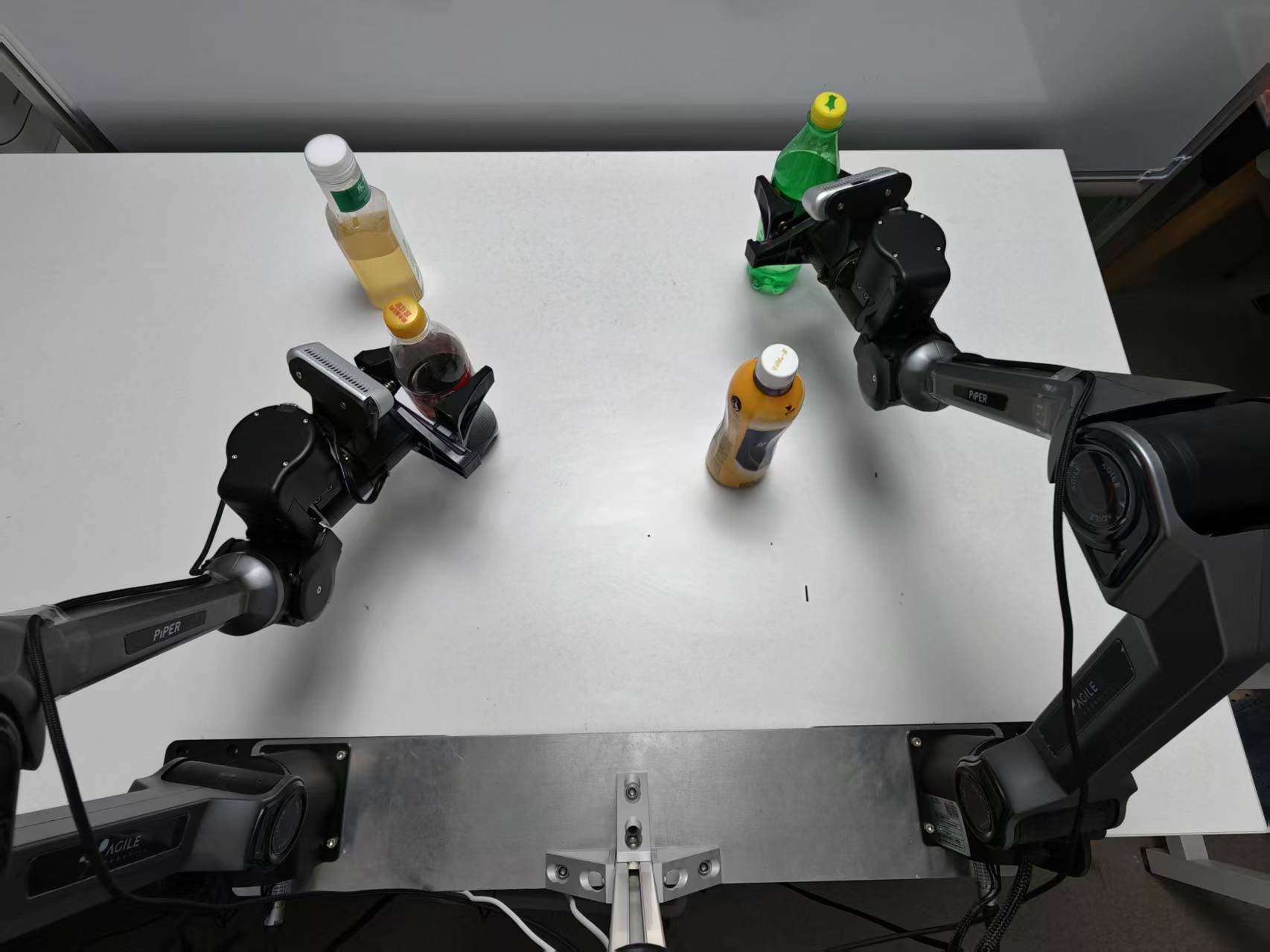} \\[-1pt]
      (a) PiperX platform & (b) Target only & (c) With distractors
  \end{tabular*}
  \caption{Real-robot setup. (a) Dual-arm Agilex PiperX with overhead and wrist cameras. (b--c) \texttt{pick\_dual\_bottles} scenes without and with
task-irrelevant distractors.}
  \label{fig:real-robot-setup}
  \vspace{-1.0em}
\end{figure}

\paragraph{Real-robot Setup.}
We deploy SVA on a dual-arm Agilex PiperX in tabletop scenes that mirror the corresponding RoboTwin tasks. The physical suite contains five tasks, which together cover object contact, transport and placement, and coordinated bimanual manipulation. Figure~\ref{fig:real-robot-setup} shows the platform and the two scene conditions, and Figure~\ref{fig:real-robot-task-scenes} in Appendix~\ref{app:real_relevance} shows the scene layout of each task.

\paragraph{Policy and Evaluator Adaptation.}
 The proposer is $\pi_{0.5}$ fine-tuned on 500 real-robot demonstration trajectories, and the Q-model, initialized from training on simulation MCTS data, is subsequently adapted on the same trajectories. We compare direct execution of the adapted $\pi_{0.5}$ with $\pi_{0.5}$+SVA under the same policy checkpoint, and both the proposer and the evaluator remain frozen during evaluation. At each decision, SVA samples and scores continuous action chunks, executes the prefix of the selected chunk, and replans from the next observation, without any online simulator search. Appendix~\ref{app:real_relevance} details the adaptation and evaluation protocol.
% The proposer is $\pi_{0.5}$ fine-tuned on 500 real-robot demonstration trajectories. The Q-model is initialized from training on simulation MCTS data and subsequently adapted with the same set of real-robot trajectories. We compare direct execution of the adapted $\pi_{0.5}$ policy with $\pi_{0.5}$+SVA using the same policy checkpoint. Policy adaptation precedes this comparison; both the proposer and the adapted evaluator remain fixed during evaluation. SVA samples and scores continuous action chunks, executes a selected prefix, and replans from the next observation without online simulator search. Appendix~\ref{app:real_relevance} describes the adaptation and evaluation protocol.

\paragraph{Scene Conditions and Evaluation.}
Each task is evaluated for 25 trials per method under two conditions: \emph{target-only}, containing only the task-relevant objects, and \emph{with-distractors}, which adds task-irrelevant objects to the same workspace (Table~\ref{tab:real-robot-main}). SVA raises the average success rate from 52.8 to 63.2 (\textbf{+10.4}) in the target-only and from 32.8 to 48.0 (\textbf{+15.2}) with distractors, outperforming $\pi_{0.5}$ on most task. Distractors degrade $\pi_{0.5}$ by 20.0 points but SVA by only 15.2, so the benefit widens as the scene becomes more cluttered. This robustness can be attributed to SVA's outcome-driven selection: by ranking candidates according to their predicted consequences rather than the proposer's likelihood, it filters out chunks that engage irrelevant objects before they are executed.
% Each task is evaluated in two conditions: \emph{target only}, containing the task-relevant objects and fixtures, and \emph{with distractors}, containing the same task-relevant objects together with additional, task-irrelevant objects. Each method receives 25 trials per task in each condition. Table~\ref{tab:real-robot-main} separates the two conditions so that performance with distractors can be assessed alongside target-only performance.

\begin{table}[t]
\centering
\caption{Real-world results on the dual-arm Agilex PiperX. Task entries report success rate (\%).}
\label{tab:real-robot-main}
\small
\setlength{\tabcolsep}{3pt}
\renewcommand{\arraystretch}{1.1}
\begin{tabularx}{\textwidth}{@{}lYYYY@{}}
\toprule
& \multicolumn{2}{c}{Target only} & \multicolumn{2}{c}{With distractors} \\
\cmidrule(lr){2-3}\cmidrule(l){4-5}
Task & $\pi_{0.5}$ & $\pi_{0.5}$+SVA & $\pi_{0.5}$ & $\pi_{0.5}$+SVA \\
\midrule
\texttt{click\_bell} & 28\% [7/25] & 48\% [12/25] & 20\% [5/25] & 48\% [12/25] \\
\texttt{move\_pillbottle\_pad} & 68\% [17/25] & 88\% [22/25] & 36\% [9/25] & 40\% [10/25] \\
\texttt{pick\_dual\_bottles} & 84\% [21/25] & 88\% [22/25] & 56\% [14/25] & 76\% [19/25] \\
\texttt{place\_mouse\_pad} & 24\% [6/25] & 32\% [8/25] & 12\% [3/25] & 16\% [4/25] \\
\texttt{stack\_blocks\_two} & 60\% [15/25] & 60\% [15/25] & 40\% [10/25] & 60\% [15/25] \\
\midrule
Avg. & 52.8\% [66/125] & 63.2\% [79/125] & 32.8\% [41/125] & 48\% [60/125] \\
\bottomrule
\end{tabularx}
\end{table}

\subsection{Ablation Study}
\label{sec:ablation}

\begin{wrapfigure}{R}{0.42\textwidth}
    \centering
    \vspace{-1em}
    \includegraphics[width=0.9\linewidth]{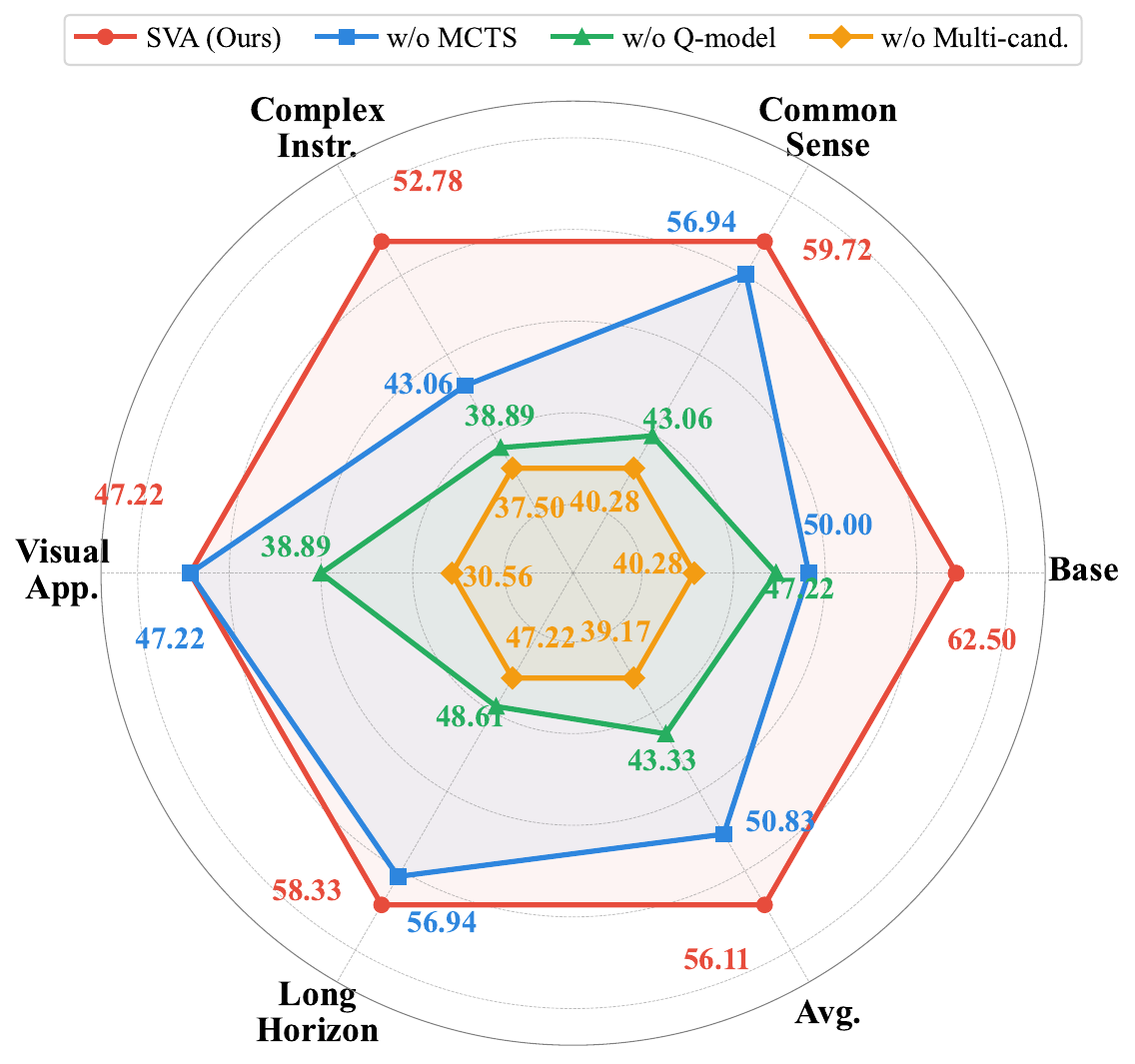}
    \caption{Component ablation on EB-Navigation.}
    \label{fig:ablation_main}
    \vspace{1.2em}
    \expandafter\def\csname @captype\endcsname{table}
    \caption{Effect of search-data size (EB-Habitat, Qwen3.5-9B).}
    \label{tab:distillation-data-size}
    \footnotesize
    \setlength{\tabcolsep}{3pt}
    \renewcommand{\arraystretch}{1.1}
    {\color{black}%
    \begin{tabular*}{\linewidth}{@{\extracolsep{\fill}}lcc}
    \toprule
    Method & Training data (\%) & Success (\%) \\
    \midrule
    Base policy & -- & 42.78 \\
    SVA & 25 & 45.42 \\
    SVA & 50 & 52.08 \\
    SVA & 100 & \textbf{57.22} \\
    \bottomrule
    \end{tabular*}}
    \vspace{-1em}
\end{wrapfigure}

\paragraph{Component Ablation.} To isolate each component's contribution, we evaluate three ablations: 1)~\textbf{w/o MCTS}, training the Q-model with only policy-rollout returns; 2)~\textbf{w/o Q-model}, replacing value evaluation with majority voting; and 3)~\textbf{w/o multi-cand.}, querying a single candidate. Results in Fig.~\ref{fig:ablation_main} show that: 
i) Dropping MCTS causes a moderate decline on EB-Navigation (56.11 vs.\ 50.83);
% while EB-Habitat stays comparable (57.22 vs.\ 58.89), indicating that MCTS targets matter most for long-horizon, sparse-reward tasks; 
ii) Removing the Q-model lowers success from 56.11\% to 43.33\% on EB-Navigation, showing the benefit of learned value estimates over majority voting;
% iii) Eliminating multi-candidate selection reduces performance to base-policy level (42.78/39.17), showing that the Q-model is effective only when used \emph{comparatively}. 
iii) Eliminating multi-candidate selection reduces performance to the single-shot backbone (56.11$\to$39.17). 
Full SVA exceeds every ablation by at least 5.28 percentage points in average success rate, supporting the contribution of each component.

\paragraph{Search-Data Scale.}
To examine how the scale of search data shapes the evaluator, we train it on a quarter, half, and all of the search data on EB-Habitat with Qwen3.5-9B. Results in Table~\ref{tab:distillation-data-size} show that: i) a quarter of the data already surpasses the base policy (42.78 to 45.42); ii) success scales monotonically with data, reaching \textbf{57.22} (\textbf{+14.44}) at full scale, with no sign of saturation within this range. Broader search coverage thus translates directly into more reliable return estimates.
% \normalcolor
% \paragraph[Distillation Data Size]{\zicanhu{Distillation Data Size.}}
% \zicanhu{On EB-Habitat with Qwen3.5-9B, using 25\% of the search data to train the evaluator raises success from 42.78\% to 45.42\%. Success increases with the data fraction, reaching 57.22\% with the full dataset (Table~\ref{tab:distillation-data-size}).}

% \subsection{Effect of Candidate Number and Deployment Efficiency}
% \subsection{Does Scaling Evaluation with SVA Yield Better Pareto Efficiency Than Scaling Parameters?}

\subsection{Scaling SVA at Test Time: Performance, Latency, and Cost-Effectiveness}
\label{sec:candidate-latency}

\begin{wrapfigure}[14]{R}{0.45\textwidth}\centering
    \vspace{-0.8em}
    \includegraphics[width=1.0\linewidth]{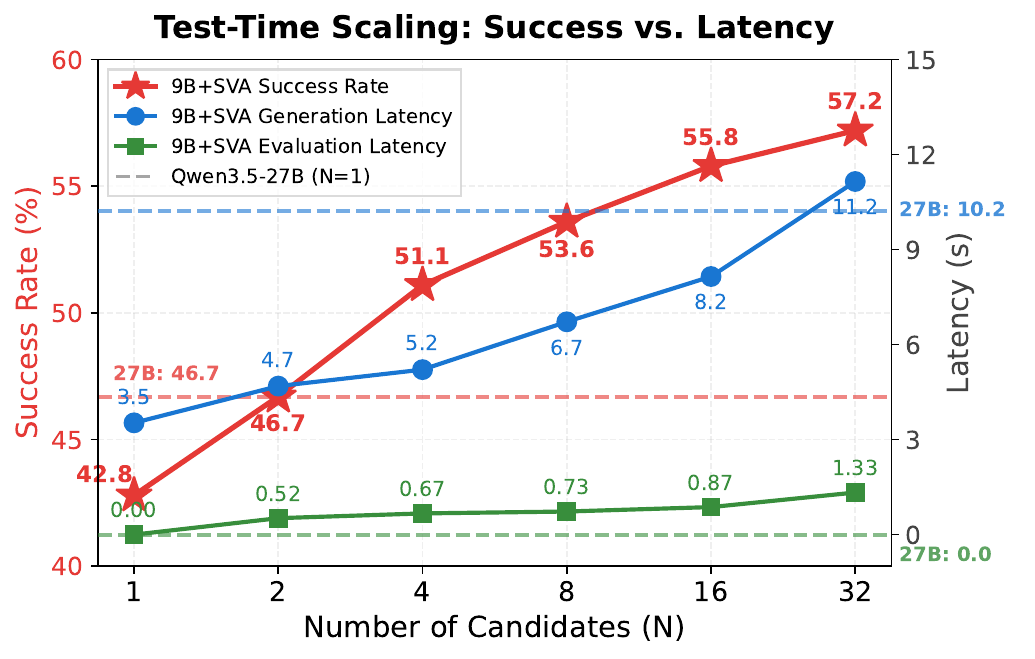}
    \caption{Success and latency versus candidate budget $N$ for Qwen3.5-9B with SVA.}
    \label{fig:latency}
    \vspace{-0.8em}
\end{wrapfigure}

\textbf{SVA exhibits strong test-time scaling.}
We analyze SVA's scaling behavior by sweeping the number of candidate actions $N$ at deployment and measuring both success rate and inference latency. 
As shown in Fig.~\ref{fig:latency}, increasing $N$ yields monotonic gains in measured success for SVA (9B) over the tested candidate budgets. 
This high-level reasoning experiment measures how additional candidate generation combines with a fixed evaluator to improve action selection.

\textbf{SVA's inference latency grows sub-linearly in $N$.}
For high-level reasoning, the evaluation latency remains well below the generation latency ($\leq 1.33$s vs.\ up to $11.2$s at $N{=}32$), and generation itself scales sub-linearly: each doubling of the candidate set ($1{\to}2, \ldots, 16{\to}32$) increases latency by only $10.63\%$--$36.58\%$ ($26\%$ on average). The same holds for continuous control: for $\pi_{0.5}$, increasing $N=K$ from 1 to 64 raises the end-to-end Act latency only from 120.1 to 340.9\,ms, a $2.84\times$ increase for a $64\times$ larger candidate set, enabled by shared-prefix sampling and CUDA Graph Q scoring. Implementation details of the accelerated pipeline are given in Appendix~\ref{app:pi05-act-latency}.

\noindent\textbf{Scaling test-time evaluation is more cost-effective than scaling model size.} 
Single-shot Qwen3.5-27B reaches $46.7\%$ success at $10.2$s of inference latency. 
In contrast, Qwen3.5-9B\,+\,SVA reaches 51.1\% at 5.87s with best-of-4 and 53.6\% at 7.43s with best-of-8: \textbf{a 7-point gain over the 27B baseline at 27\% lower latency}, using a $3\times$ smaller backbone. 
This comparison supports candidate evaluation as an alternative allocation of inference compute in the measured setting, complementing test-time scaling findings in LLMs~\citep{snell2025scaling}.

% \paragraph{continuous control Act Latency.}
% With shared-prefix sampling and CUDA Graph Q scoring, increasing $N=K$ from 1 to 64 raises the measured $\pi_{0.5}$ Act end-to-end mean latency from $120.119$ to $340.947$\,ms ($2.84\times$). Appendix~\ref{app:pi05-act-latency} describes the accelerated path and reports the component-wise latency curves in Fig.~\ref{fig:pi05-act-latency}.

% \begin{table}[t]
% \centering
% \caption{Effect of candidate number using Qwen3.5-9B. Success rate is reported in \%; latency is reported in seconds.}
% \label{tab:latency}
% \setlength{\tabcolsep}{4pt}
% \begin{tabular}{lcccccc}
% \toprule
% \# Candidates & 1 & 2 & 4 & 8 & 16 & 32 \\
% \midrule
% Success rate & 42.8 & 46.2 & 52.9 & 53.3 & 55.8 & 57.5 \\
% Action proposal latency & 3.54 & 4.77 & 5.14 & 6.65 & 8.40 & 11.05 \\
% Action scoring latency & 0.00 & 0.56 & 0.74 & 0.78 & 0.86 & 1.34 \\
% \bottomrule
% \end{tabular}%
% \end{table}

% \subsection{Ablation Study}\label{subsec:ablation}
% We ablate the contribution of each SVA component on Qwen3.5-9B. Full numerical results are provided in Table~\ref{tab:ablation} in the Appendix.~\ref{app:more_results}.

% \subsection{Can SVA achieve inference latency that grows sub-linearly in $N$?}\label{subsec:lantency}

% \subsection{Can scale test-time evaluation in SVA be more cost-effective than scaling model size?}\label{subsec:scale}

\subsection{Case Study}
\label{sec:case_study}
\paragraph{Embodied Reasoning: Overriding Distractors.} We illustrate SVA with the instruction \emph{On the sofa there's an apple, but instead find a plate and move it to the brown table.} The first clause introduces a salient but irrelevant apple--sofa pair. Full results are in Appendix~\ref{app:case_study}. As shown in Fig.~\ref{fig:case1-study}, the base policy follows the distractor to the sofa and then issues repeated invalid actions, whereas SVA selects a plate-centric plan and reaches the target table in 4 steps without invalid actions. Table~\ref{tab:case-study-ranking} explains the difference: the proposer samples the sofa-first plan more often, but the Q-network values the plate-centric plan higher. Long-horizon value estimates thus override myopic proposal preferences.
% We provide a qualitative case study with the instruction: \emph{On the sofa there's an apple, but instead find a plate and move it to the brown table.} This task is challenging because the first clause introduces a salient but irrelevant object-location pair (apple-sofa), distracting from the true goal of moving the plate to the brown table.
% Full results are provided in Appendix~\ref{app:case_study}.
% As shown in Fig.~\ref{fig:case1-study}, the base policy is misled by the distractor, navigating to the sofa and then issuing repeated invalid recovery actions. 

% \begin{table}[t]
%     \centering
%     \caption{Candidate ranking at the first decision step. Although the goal-directed plan has a lower proposal prior, the learned Q-network assigns it a higher value and selects it for execution.}
%     \label{tab:case-study-ranking}
%     \setlength{\tabcolsep}{5pt}
%     \renewcommand{\arraystretch}{1.15}
%     \begin{tabular}{lccc}
%         \toprule
%         Candidate plan & Proposal prior & Q-value & Selected \\
%         \midrule
%         \makecell[l]{Distractor-driven plan: \\ navigate to the sofa/apple} 
%         & \textbf{0.4375} & -0.1282 & \textcolor{baseRed}{\xmark} \\
%         \rowcolor{lightBlue}
%         \makecell[l]{Goal-directed plan: \\ find plate and move to brown table} 
%         & 0.0625 & \textbf{0.2948} & \textcolor{svaBlue}{\cmark} \\
%         \bottomrule
%     \end{tabular}%
% \end{table}
\begin{wraptable}{r}{0.43\textwidth}
    \vspace{-1.5em}
    \centering
    \small
    \caption{
    % Candidate ranking at the first step. Lower prior but higher Q-value leads to correct selection.
    Action evaluation: lower prior but higher Q-value leads to correct selection.
    }
    \label{tab:case-study-ranking}
    \setlength{\tabcolsep}{2pt}
    \renewcommand{\arraystretch}{1.1}
    \begin{tabular}{lccc}
        \toprule
        \textbf{Candidate} & \textbf{Prior} & \textbf{Q-value} & \textbf{Selected} \\
        \midrule
        Distractor-driven
        & \textbf{0.4375} & -0.1282 & \textcolor{baseRed}{\xmark} \\
        \rowcolor{lightBlue}
        Goal-directed
        & 0.0625 & \textbf{0.2948} & \textcolor{svaBlue}{\cmark} \\
        \bottomrule
    \end{tabular}%
    \vspace{-0.8em}
\end{wraptable}

% In contrast, SVA selects a plate-centric plan and reaches the target table in just 4 steps with no invalid actions.
% The candidate scores in Table~\ref{tab:case-study-ranking} illustrate the selection: the distractor-driven sofa-first plan has a higher empirical sampling frequency, while the Q-network assigns a higher value to the plate-centric plan.
% The value model can override myopic proposal preferences via long-horizon consequence evaluation, improving robustness and reducing invalid behaviors.

\begin{figure}[tb]
      \centering
      \footnotesize
      \setlength{\fboxsep}{0pt}
      \setlength{\fboxrule}{0.6pt}
      \setlength{\tabcolsep}{0pt}
      % Frames are 640x480; the empty tabletop at the bottom is trimmed (trim order: left, bottom, right, top).
      \begin{tabular*}{\textwidth}{@{\extracolsep{\fill}}ccccc@{}}
          & Step 0 & Step 24 & Step 48 & Step 72 \\[1pt]
          \parbox[c]{1.2em}{\rotatebox{90}{\textcolor{baseRed}{$\pi_{0.5}$~\xmark}}}
          & \parbox[c]{0.232\textwidth}{\includegraphics[width=0.232\textwidth,trim={0bp 140bp 0bp
  0bp},clip]{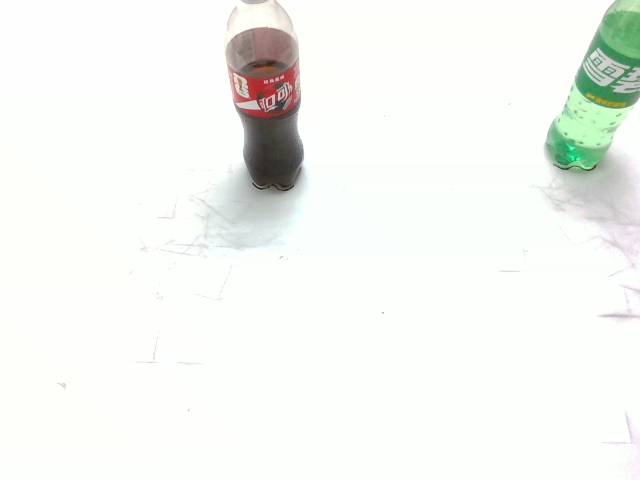}}
          & \parbox[c]{0.232\textwidth}{\includegraphics[width=0.232\textwidth,trim={0bp 140bp 0bp
  0bp},clip]{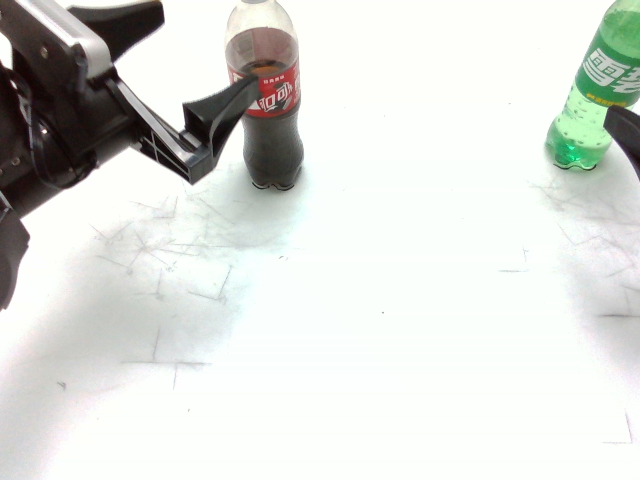}}
          & \parbox[c]{0.232\textwidth}{\includegraphics[width=0.232\textwidth,trim={0bp 140bp 0bp
  0bp},clip]{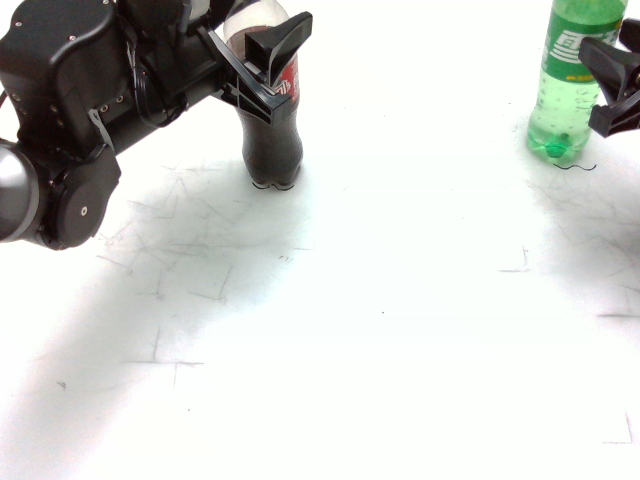}}
          & \parbox[c]{0.232\textwidth}{\includegraphics[width=0.232\textwidth,trim={0bp 140bp 0bp
  0bp},clip]{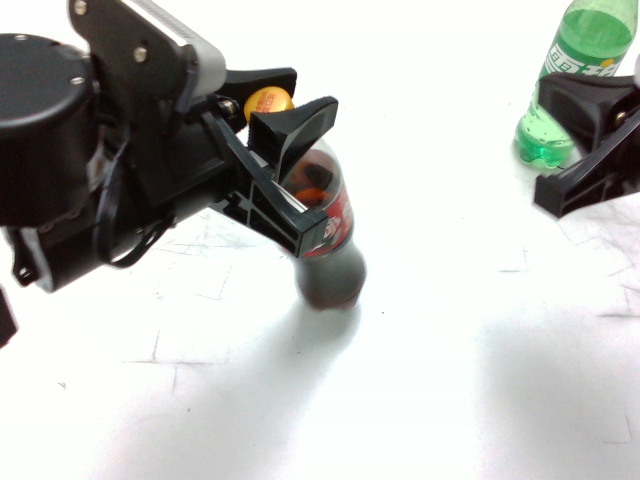}} \\[2pt]
          \parbox[c]{1.2em}{\rotatebox{90}{\textcolor{svaBlue}{$\pi_{0.5}$+SVA~\cmark}}}
          & \parbox[c]{0.232\textwidth}{\includegraphics[width=0.232\textwidth,trim={0bp 140bp 0bp
  0bp},clip]{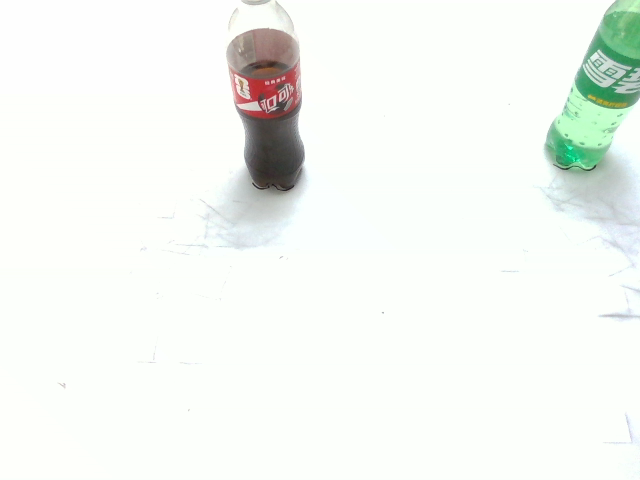}}
          & \parbox[c]{0.232\textwidth}{\includegraphics[width=0.232\textwidth,trim={0bp 140bp 0bp
  0bp},clip]{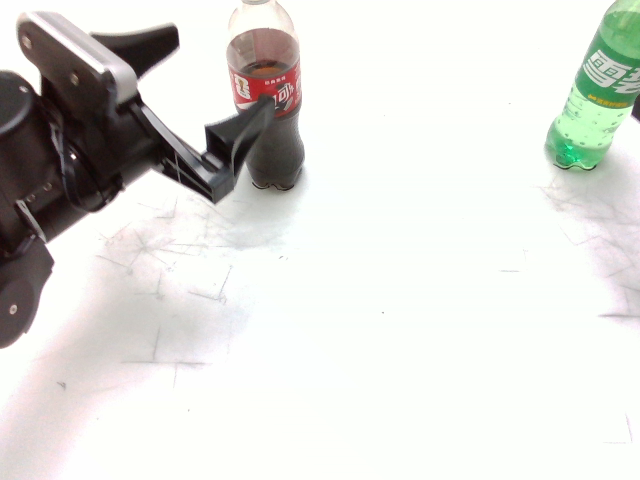}}
          & \parbox[c]{0.232\textwidth}{\fcolorbox{svaBlue}{white}{\includegraphics[width=\dimexpr0.232\textwidth-2\fboxrule\relax,trim={0bp 140bp 0bp
  0bp},clip]{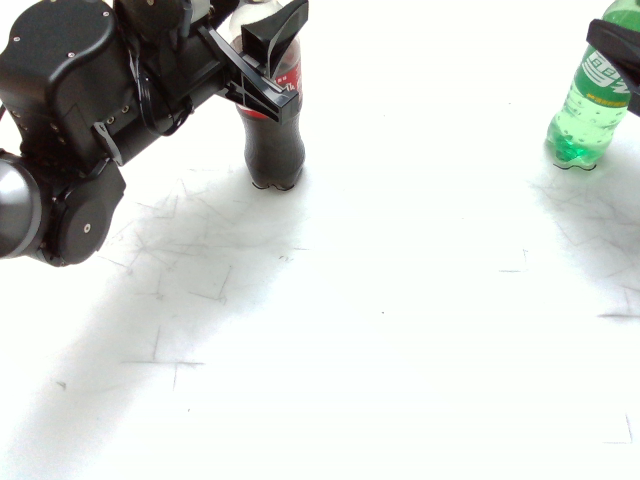}}}
          & \parbox[c]{0.232\textwidth}{\fcolorbox{svaBlue}{white}{\includegraphics[width=\dimexpr0.232\textwidth-2\fboxrule\relax,trim={0bp 140bp 0bp
  0bp},clip]{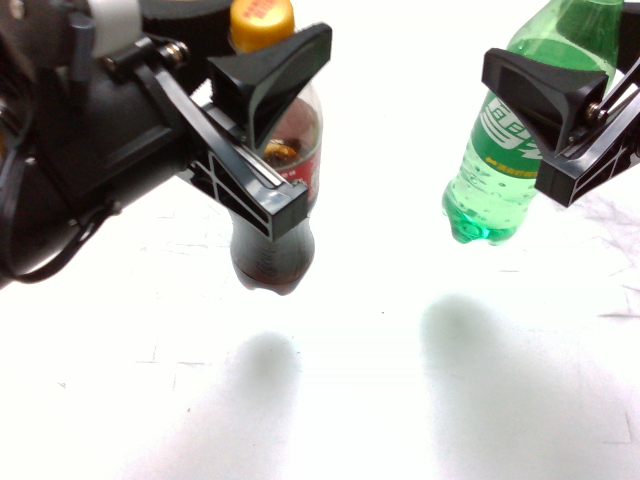}}}
      \end{tabular*}
      \caption{Real-world \texttt{pick\_dual\_bottles} case at matched execution steps. Boxed SVA frames show the observation where candidate \#1 is
  selected and the outcome after executing it.}
      \label{fig:real-robot-case}
  \end{figure}

\paragraph{Real World: From Selection to Success.}
Figure~\ref{fig:real-robot-case} compares $\pi_{0.5}$ and $\pi_{0.5}$+SVA on \texttt{pick\_dual\_bottles} under the same instruction and similar initial layouts. At step 48, SVA scores 16 distinct action chunks and selects candidate \#1, which has the highest mean Q-value (0.80). After this chunk executes over steps 49--72, both bottles are lifted, whereas the baseline's green bottle remains on the table at the same step. The baseline eventually reaches the 200-step limit without success, while SVA succeeds at step 95. This case links a single Q-based selection directly to the subsequent
  physical outcome.
% Figure~\ref{fig:real-robot-case} compares $\pi_{0.5}$ and $\pi_{0.5}$+SVA on \texttt{pick\_dual\_bottles} with the same instruction and similar initial layouts. At step 48, SVA scores 16 distinct action chunks and selects candidate \#1 (zero-based), whose mean Q-value of $0.80094$ is the highest. The selected chunk passes through the controller and executes over steps 49--72. The next observation shows both bottles lifted. At the same observation step, the baseline's green bottle remains on the table. The baseline ultimately reaches the 200-step limit without success, whereas SVA receives operator-confirmed success at step 95. This case connects a recorded Q-based selection to the subsequent physical outcome.

% \begin{wrapfigure}{r}{0.52\textwidth}
%     \centering
%     \includegraphics[width=1.0\linewidth]{Figs/ablation_navigation.pdf}
%     \caption{Ablation study on EB-Navigation (Qwen3.5-9B). SVA (Ours) consistently achieves the highest success rates across most task categories. See Fig.~\ref{fig:ablation_study} in the Appendix for full results.}
%     \label{fig:ablation_navigation}
% \end{wrapfigure}

% \subsection{Analysis}\label{subsec:analysic}\paragraph{}

%=====================================================
\section{Discussion and Limitations}
% We studied search-derived return supervision for selecting among candidates from a fixed embodied policy, covering high-level atomic skill sequences and continuous robot action chunks.
% Inspired by this, we proposed SVA, a search-and-learning recipe that distills MCTS-derived long-horizon consequence into a lightweight Q-model and uses it to score candidate actions at deployment without simulator access. 
% Across the evaluated benchmarks, SVA improves average success with the tested proposers; the high-level latency study shows a favorable success--latency trade-off relative to the compared larger model.
% We hope these results encourage the community to view evaluation as a first-class lever for VLA improvement, orthogonal to data scaling and policy fine-tuning.
We presented SVA, which distills simulator-backed search into a value model that reranks candidates from a frozen embodied policy, for both high-level skill sequences and continuous action chunks. Across simulated and real-world benchmarks, SVA improves average success over every tested proposer, and a 9B policy with SVA outperforms a 27B policy at lower latency. These results position action evaluation as a first-class lever for VLAs, complementary to data scaling and policy fine-tuning.

% \zicanhu{Remaining challenges include coupling search with value learning, relaxing the offline simulator requirement, and assessing robustness under sim-to-real shift.} 
% SVA is limited by the quality of search supervision and the coverage of policy-generated candidates. Its staged pipeline prevents search and the learned evaluator from improving each other; future work could close this loop and cautiously adapt the policy. Search also assumes a resettable simulator with task rewards, which could be relaxed using learned world models or sparse outcome labels. Finally, reducing real-world adaptation data and improving transfer across tasks and hardware remain important directions.
% Several directions could further extend SVA. Its effectiveness naturally depends on the quality of search supervision and the diversity of policy-generated candidates. The current staged pipeline could be turned into a closed loop in which search and the learned evaluator improve each other, and in which the policy itself is adapted in a controlled manner. Search supervision could also be extended beyond simulation, for example through learned world models or sparse outcome labels. Finally, reducing the real-world adaptation data and strengthening transfer across tasks and hardware are promising avenues for broader deployment.
SVA's gains depend on the quality of search supervision and the diversity of policy-generated candidates. Future work could close the loop between search and the evaluator, cautiously adapt the policy, and extend search supervision beyond simulation via learned world models or sparse outcome labels. Reducing real-world adaptation data and improving transfer across tasks and hardware are further directions.
See Appendix~\ref{app:limitations} for details of limitations that we leave for future work.

\clearpage
\bibliography{reference}

\clearpage
\appendix

\startcontents[appendix]
\printcontents[appendix]{}{1}{}

%=========================================================
\newpage
\section{Limitations and Future Work}\label{app:limitations}
SVA depends on the quality of search supervision and the coverage of policy-generated candidates. These dependencies leave several directions for further study.

First, our pipeline is deliberately staged with separate tree search and value learning, so search is blind to the evaluator being learned and the policy never benefits from improved values beyond reranking at test time, exploiting only a slice of what RL has to offer.
A natural next step is to unify the stages into an online loop for search and learning, where an up-to-date Q-model guides MCTS and on-policy MCTS rollouts continually refine the Q-model, unlocking more of RL's potential while carefully adapting the backbone to internalize these gains without sacrificing generalization.

Second, the Search stage relies on a resettable simulator with a task success signal; while this is standard for benchmarks such as EmbodiedBench and RoboTwin, it limits direct applicability to settings where high fidelity simulation or reward functions are unavailable.
A promising path forward is to replace, or supplement, simulator rollouts with learned world models or sparse human or automatically labeled outcomes, so that the same Search--Value--Act framework  can mine evaluation signals with improved sample efficiency in domains where ground truth simulation is impractical.

Third, our evaluation on physical robots adapts the evaluator with real trajectories to address shifts in observations and dynamics. Adaptation data requirements and transfer to new tasks or hardware need further study.

Taken together, these limitations chart a coherent research agenda: tightening the loop between search and learning, relaxing the simulator assumption, and bridging the gap between simulation and physical robots.
We view SVA as an initial step toward scalable reasoning at test time for embodied agents, and we hope the Search--Value--Act paradigm provides a foundation upon which the community can build more general, scalable, deployable, and self-improving embodied systems.

%================================================
%================================================
\section{Benchmarks}\label{app:benchmark}
Our simulation experiments use EmbodiedBench for high-level atomic skill selection, and RoboTwin 2.0 and SimplerEnv for continuous action chunk selection.
% Our main simulation experiments use EmbodiedBench for high-level atomic skill selection and RoboTwin 2.0 for continuous action chunk selection. Supplementary manipulation results on SimplerEnv are reported in Appendix~\ref{app:vla_results}.

\paragraph{EmbodiedBench.}
We use two EmbodiedBench suites. \textbf{EB-Habitat} is a household rearrangement benchmark built on Habitat. The agent receives a first person RGB observation and a natural language instruction, and executes discrete skills such as navigation, pick, place, open, and close. We evaluate the \texttt{base}, \texttt{common\_sense}, \texttt{complex\_instruction}, \texttt{spatial\_relationship}, \texttt{visual\_appearance}, and \texttt{long\_horizon} splits, with a maximum episode length of 30 environment steps. \textbf{EB-Navigation} is an object goal navigation benchmark built on AI2-THOR. Its action space contains eight primitive actions: move forward, move backward, move right, move left, rotate right, rotate left, look up, and look down. We evaluate the \texttt{base}, \texttt{common\_sense}, \texttt{complex\_instruction}, \texttt{visual\_appearance}, and \texttt{long\_horizon} splits, with a maximum episode length of 20 environment steps.

\paragraph{RoboTwin continuous control.}
Our main comparison for continuous control uses 12 tasks from the SAPIEN based RoboTwin 2.0 benchmark. Table~\ref{tab:robotwin-main} lists the complete task set and compares $\pi_{0.5}$ and FastWAM, their SVA variants, and two additional selectors based on $\pi_{0.5}$. The tasks cover both short and long manipulation sequences, as well as single arm and bimanual behaviors. The policies receive RGB observations from multiple views, the instruction, and proprioceptive information, and produce continuous action chunks. Success is determined by the simulator's task specific success check.

\paragraph{SimplerEnv.}
The SimplerEnv comparison uses four WidowX tasks from ManiSkill2: \texttt{widowx\_carrot\_on\_plate}, \texttt{widowx\_stack\_cube}, \texttt{widowx\_spoon\_on\_towel}, and \texttt{widowx\_put\_eggplant\_in\_basket}. The inputs include a front RGB image, a language instruction, and 7-D Bridge end-effector proprioception.

\begin{table}[t]
\centering
\caption{Simulation benchmark coverage.}
\small
\setlength{\tabcolsep}{3pt}
\renewcommand{\arraystretch}{1.1}
\begin{tabularx}{\textwidth}{@{}lYY@{}}
\toprule
Benchmark & Action interface & Tasks / splits \\
\midrule
EB-Habitat & Atomic skills & 6 splits \\
EB-Navigation & 8 atomic actions & 5 splits \\
RoboTwin 2.0 & Continuous chunks & 12 tasks \\
SimplerEnv & 7-D WidowX chunks & 4 tasks \\
\bottomrule
\end{tabularx}
\label{tab:app_benchmarks}
\end{table}

%========================================================
\section{Real-World Experimental Details}\label{app:real_relevance}

\subsection{Platform and Scene Conditions}
\begin{figure}[p]
    \centering
    \footnotesize
    % Common native crop for all ten 4032 x 3024 photographs; originals are unchanged.
    % Visible region: x=200..3832, y=200..2016 (2:1 aspect ratio). Trim order: left, bottom, right, top.
    \begin{tabular}{@{}c@{\hspace{0.03\textwidth}}c@{}}
        \textbf{Target only} & \textbf{With distractors} \\[3pt]
        \multicolumn{2}{@{}l@{}}{\textbf{(a)} \texttt{click\_bell}} \\[1pt]
        \includegraphics[width=0.475\textwidth,trim={200bp 1008bp 200bp 200bp},clip]{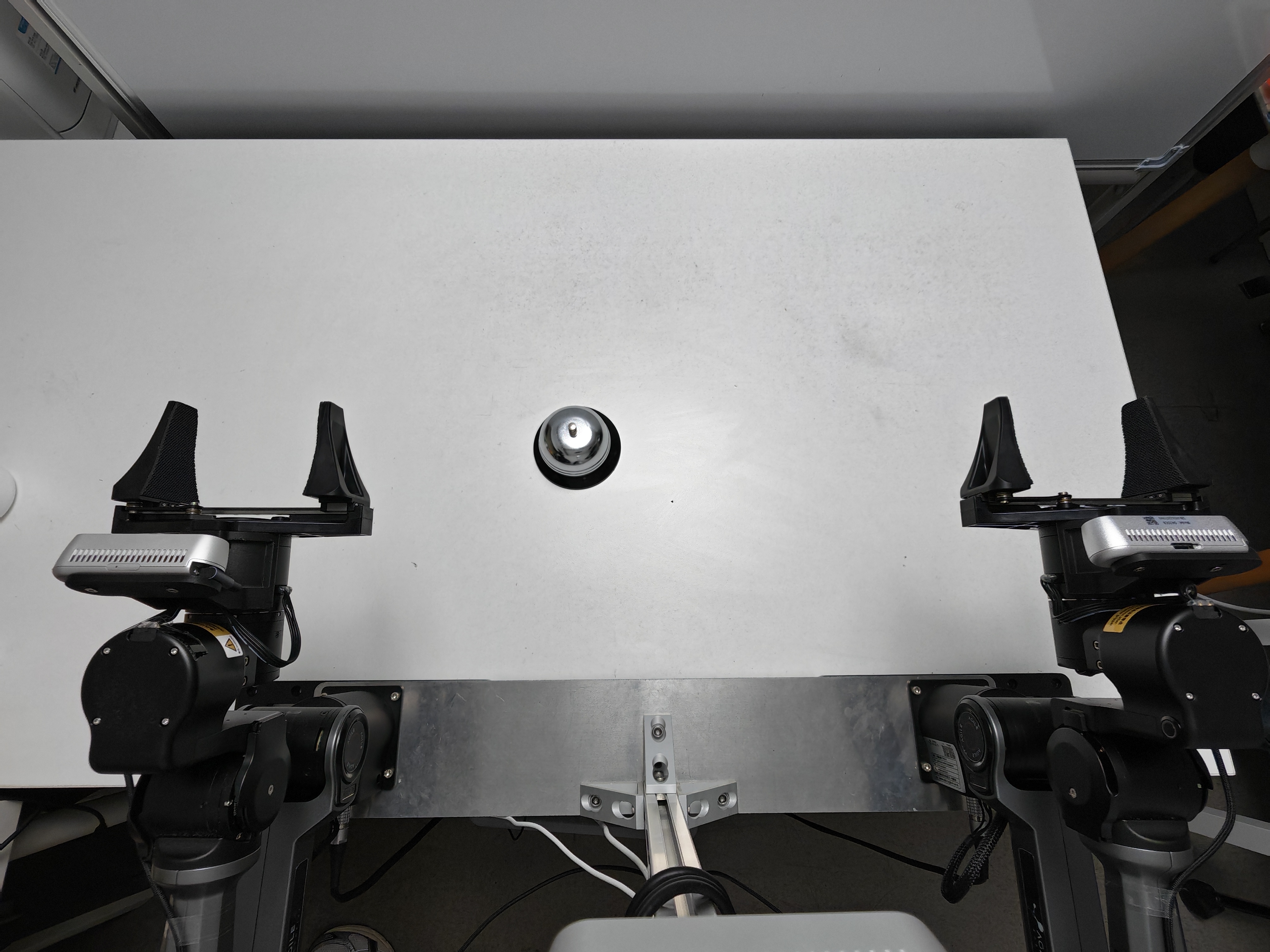}
        & \includegraphics[width=0.475\textwidth,trim={200bp 1008bp 200bp 200bp},clip]{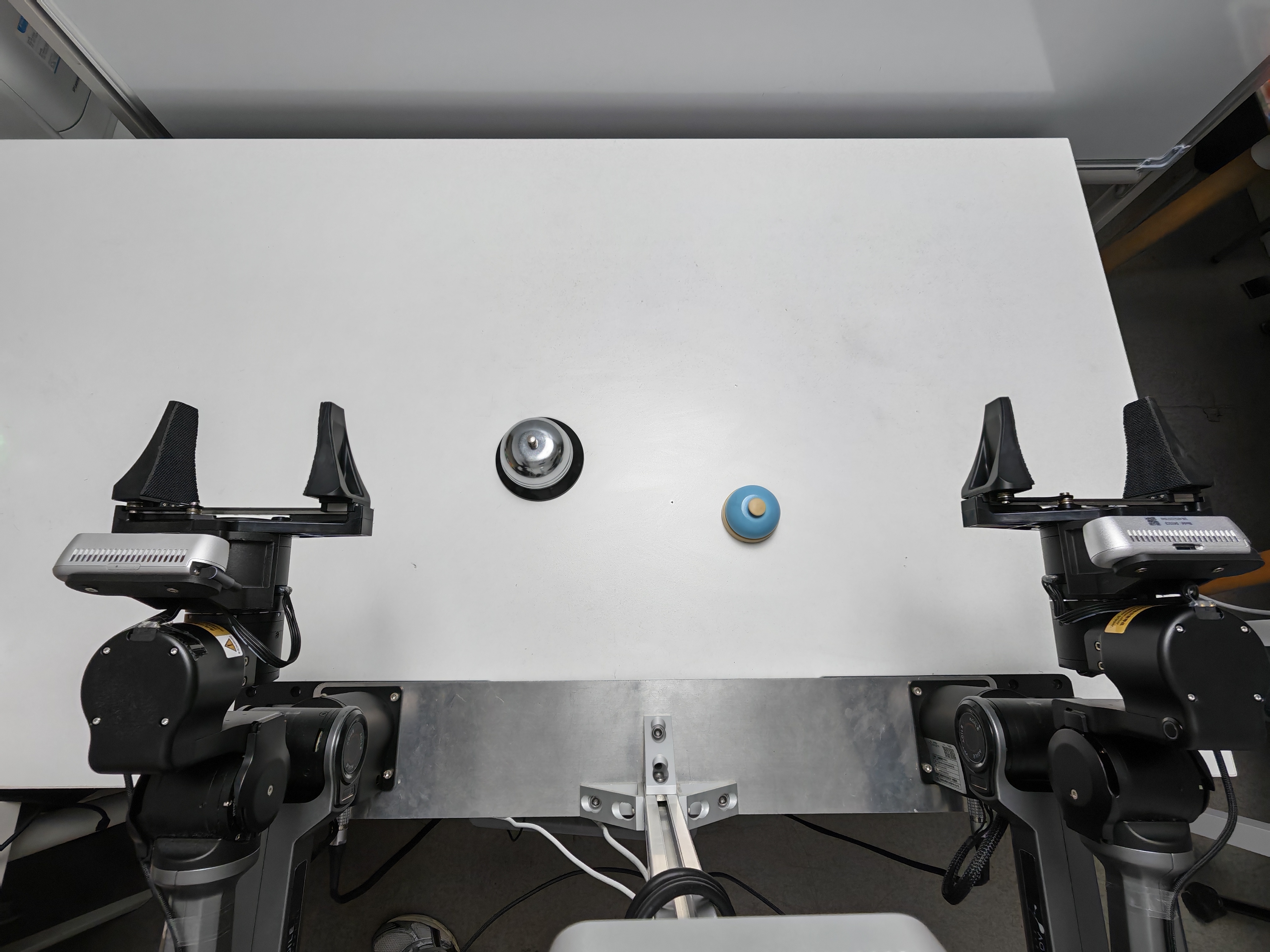} \\[4pt]
        \multicolumn{2}{@{}l@{}}{\textbf{(b)} \texttt{move\_pillbottle\_pad}} \\[1pt]
        \includegraphics[width=0.475\textwidth,trim={200bp 1008bp 200bp 200bp},clip]{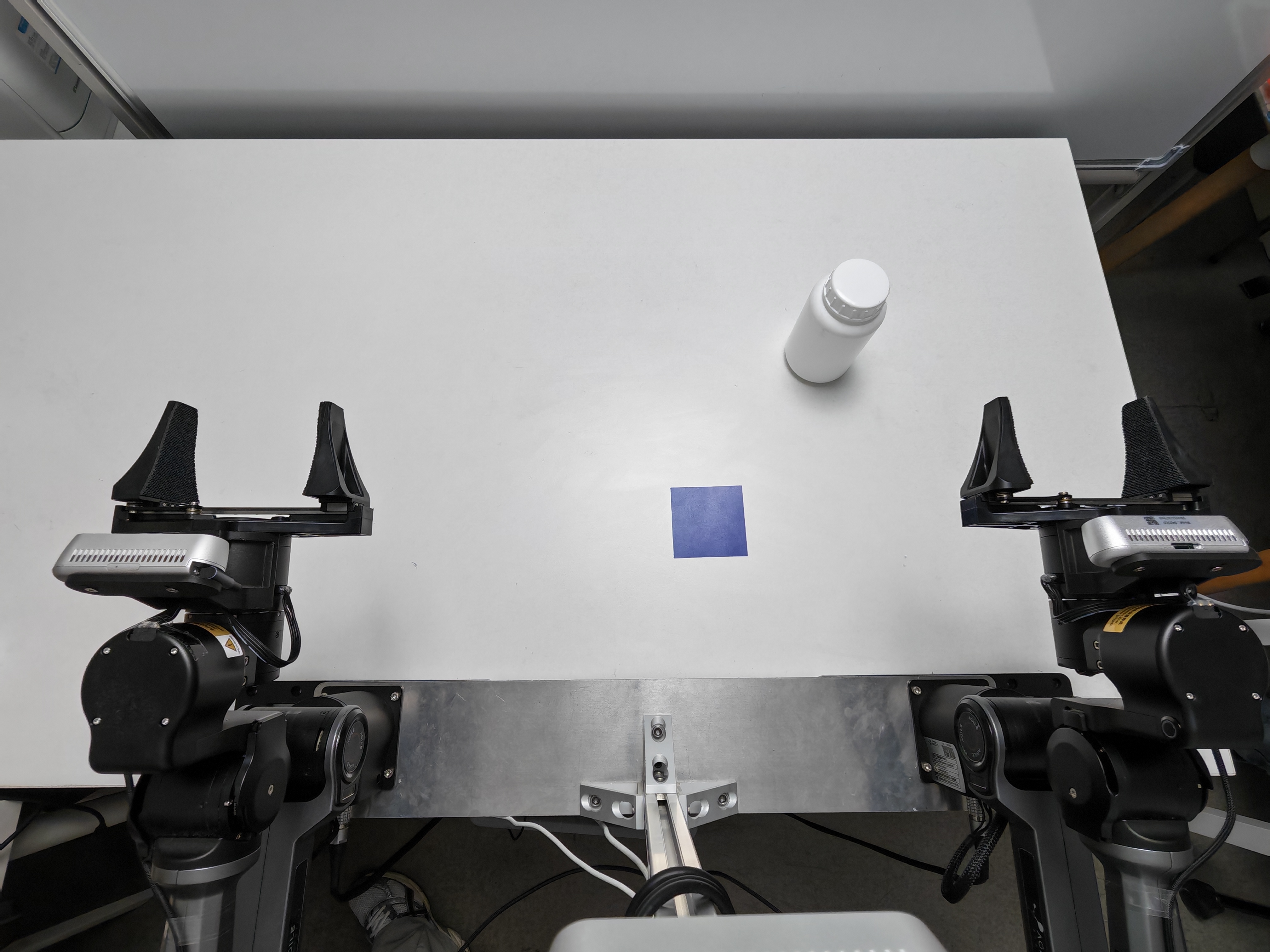}
        & \includegraphics[width=0.475\textwidth,trim={200bp 1008bp 200bp 200bp},clip]{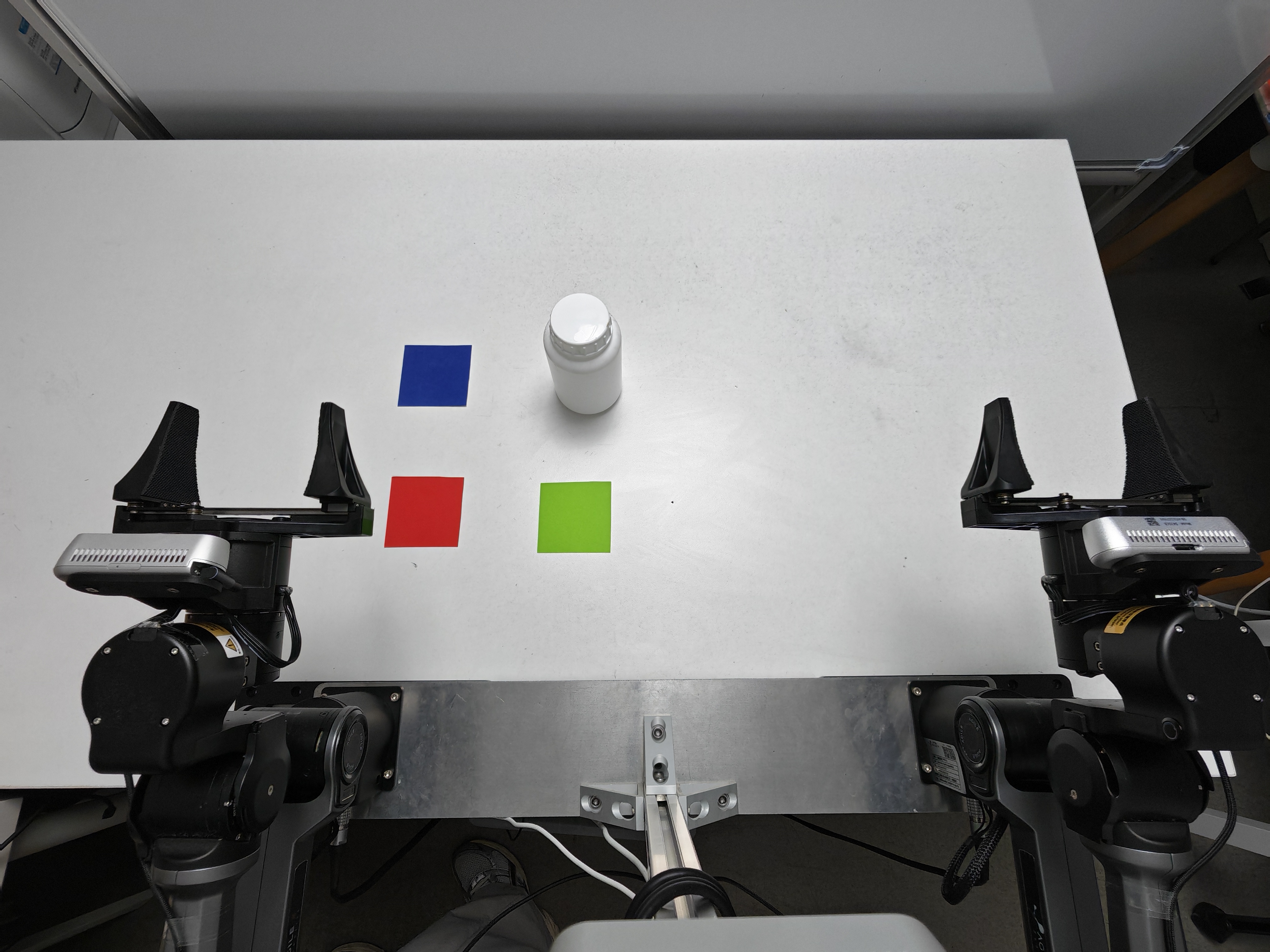} \\[4pt]
        \multicolumn{2}{@{}l@{}}{\textbf{(c)} \texttt{pick\_dual\_bottles}} \\[1pt]
        \includegraphics[width=0.475\textwidth,trim={200bp 1008bp 200bp 200bp},clip]{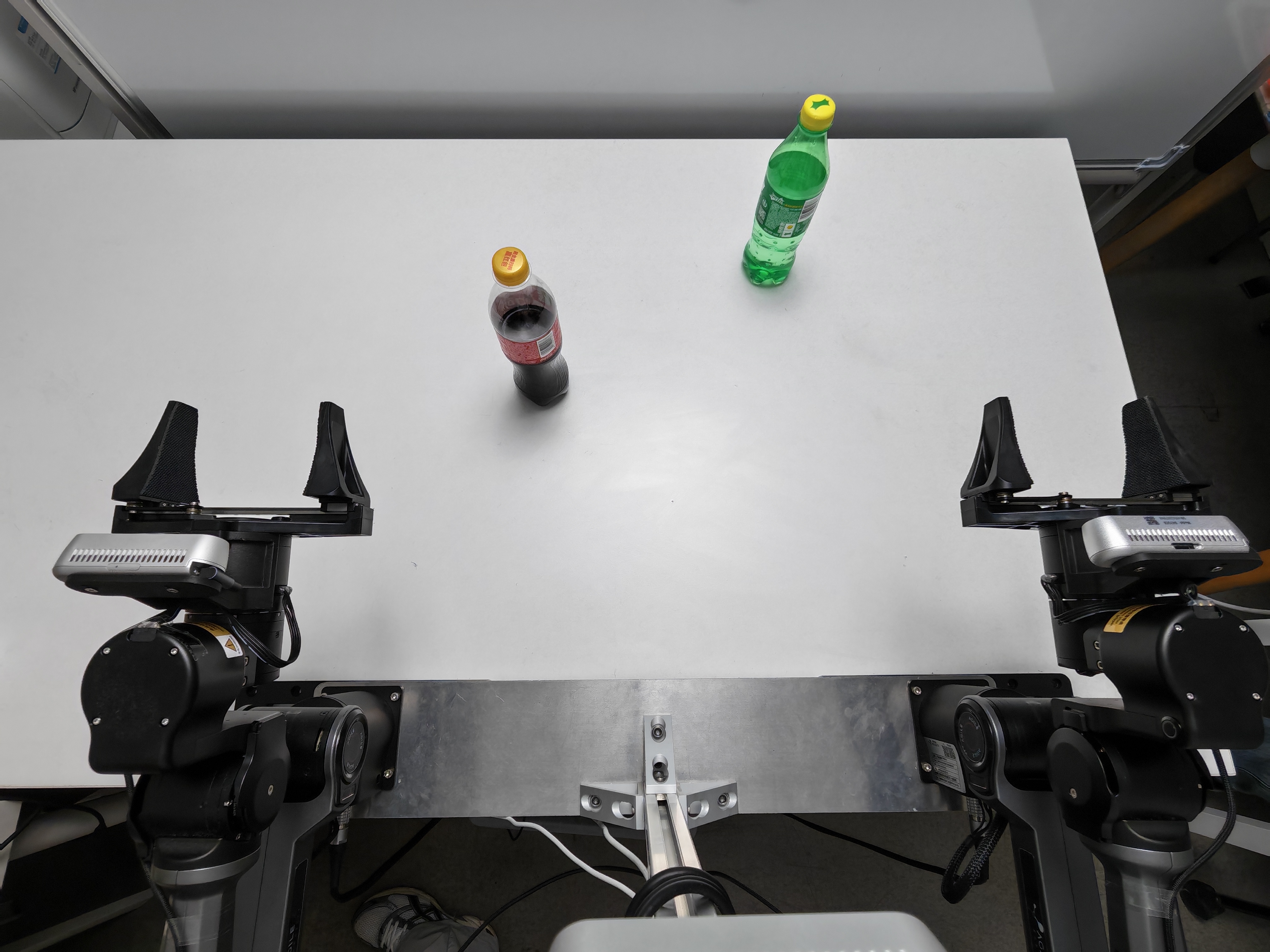}
        & \includegraphics[width=0.475\textwidth,trim={200bp 1008bp 200bp 200bp},clip]{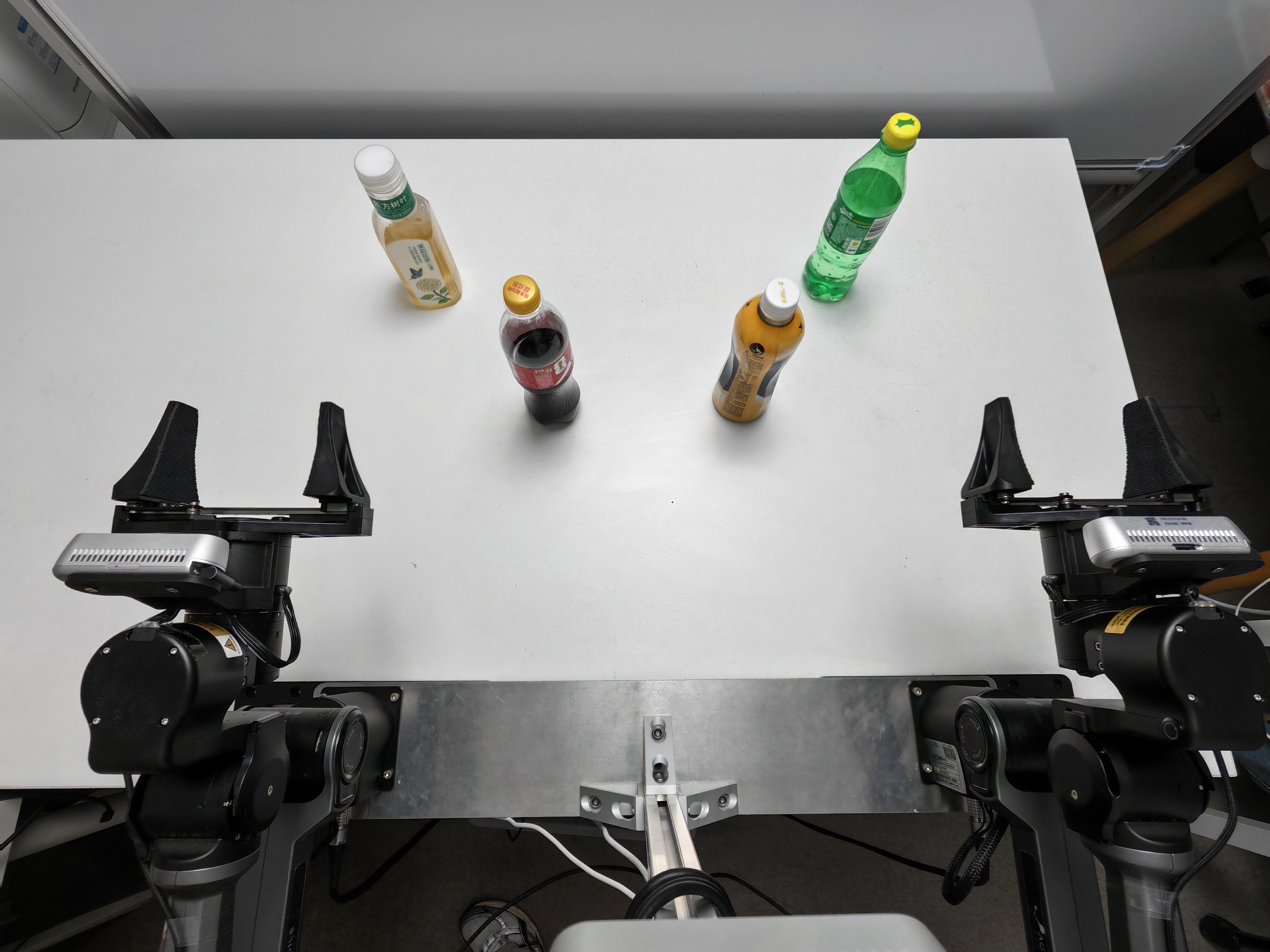} \\[4pt]
        \multicolumn{2}{@{}l@{}}{\textbf{(d)} \texttt{place\_mouse\_pad}} \\[1pt]
        \includegraphics[width=0.475\textwidth,trim={200bp 1008bp 200bp 200bp},clip]{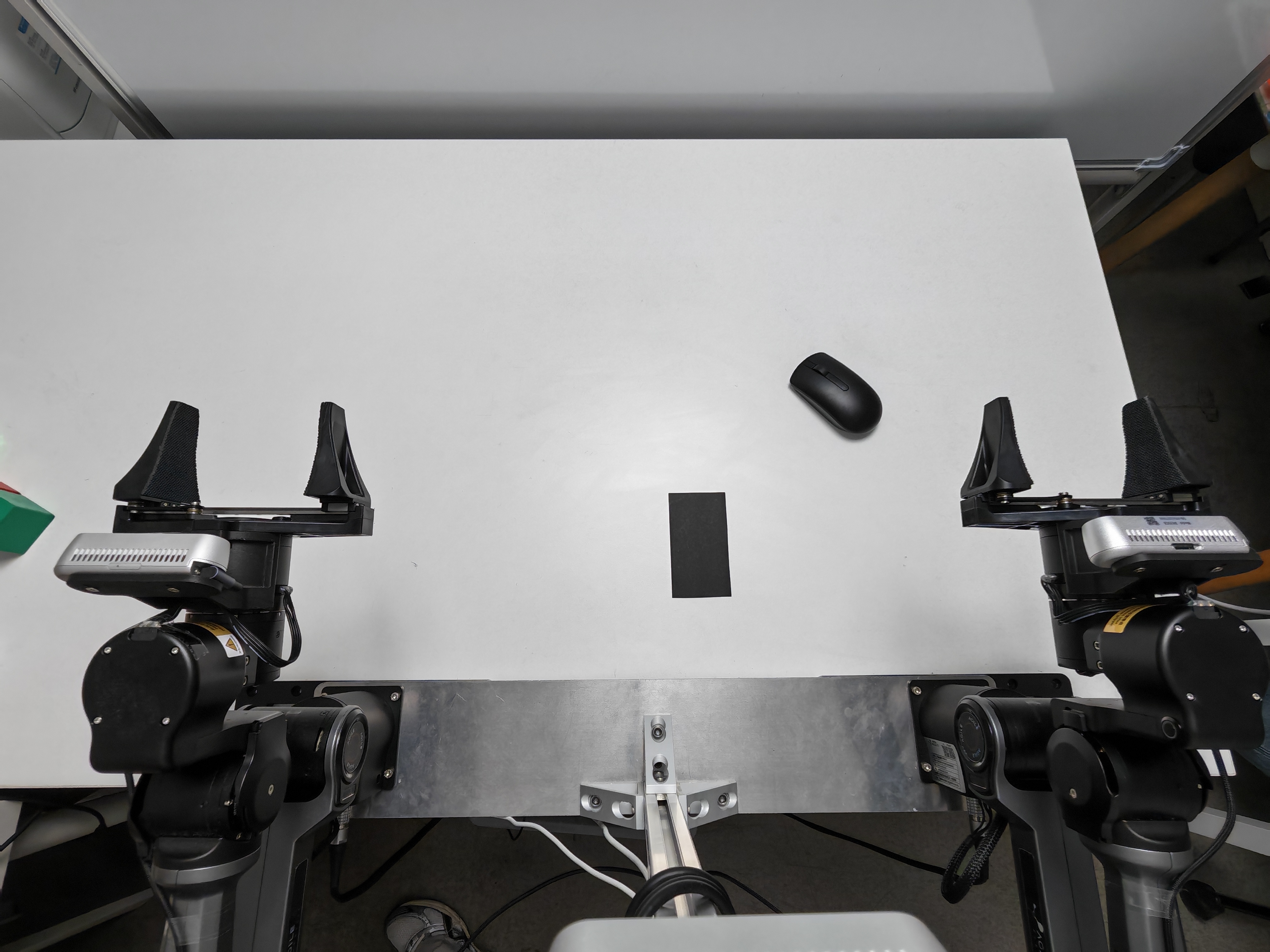}
        & \includegraphics[width=0.475\textwidth,trim={200bp 1008bp 200bp 200bp},clip]{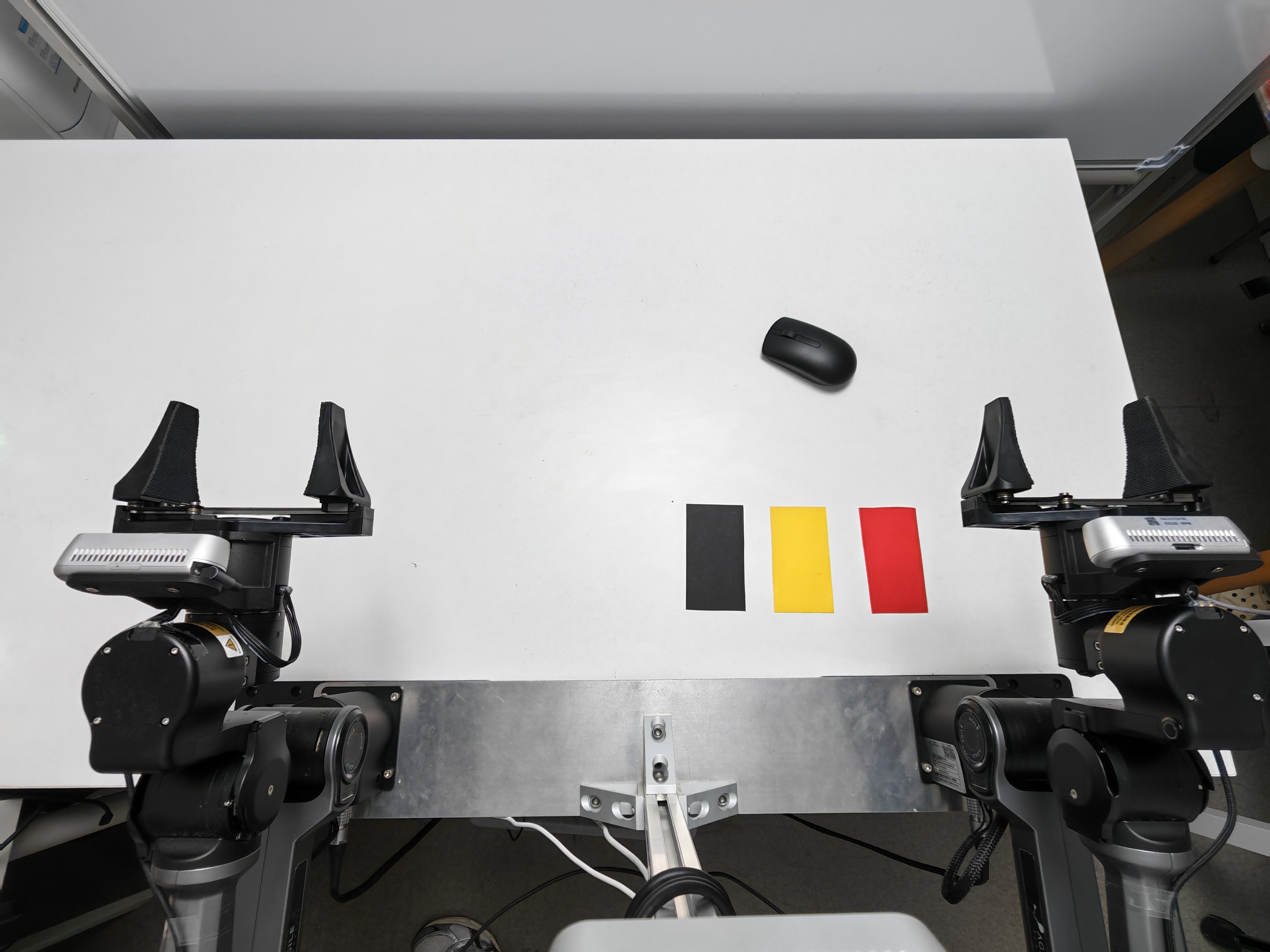} \\[4pt]
        \multicolumn{2}{@{}l@{}}{\textbf{(e)} \texttt{stack\_blocks\_two}} \\[1pt]
        \includegraphics[width=0.475\textwidth,trim={200bp 1008bp 200bp 200bp},clip]{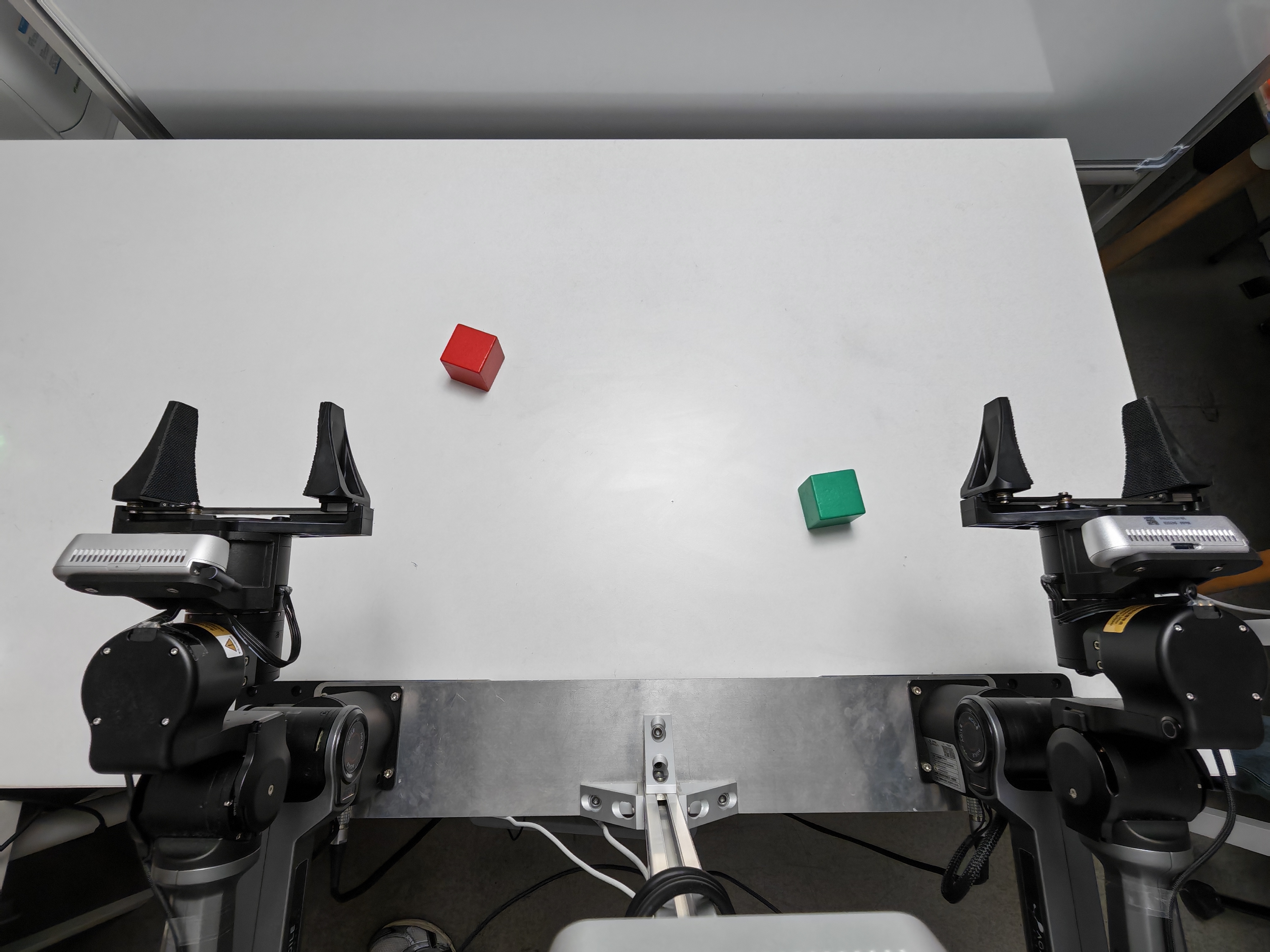}
        & \includegraphics[width=0.475\textwidth,trim={200bp 1008bp 200bp 200bp},clip]{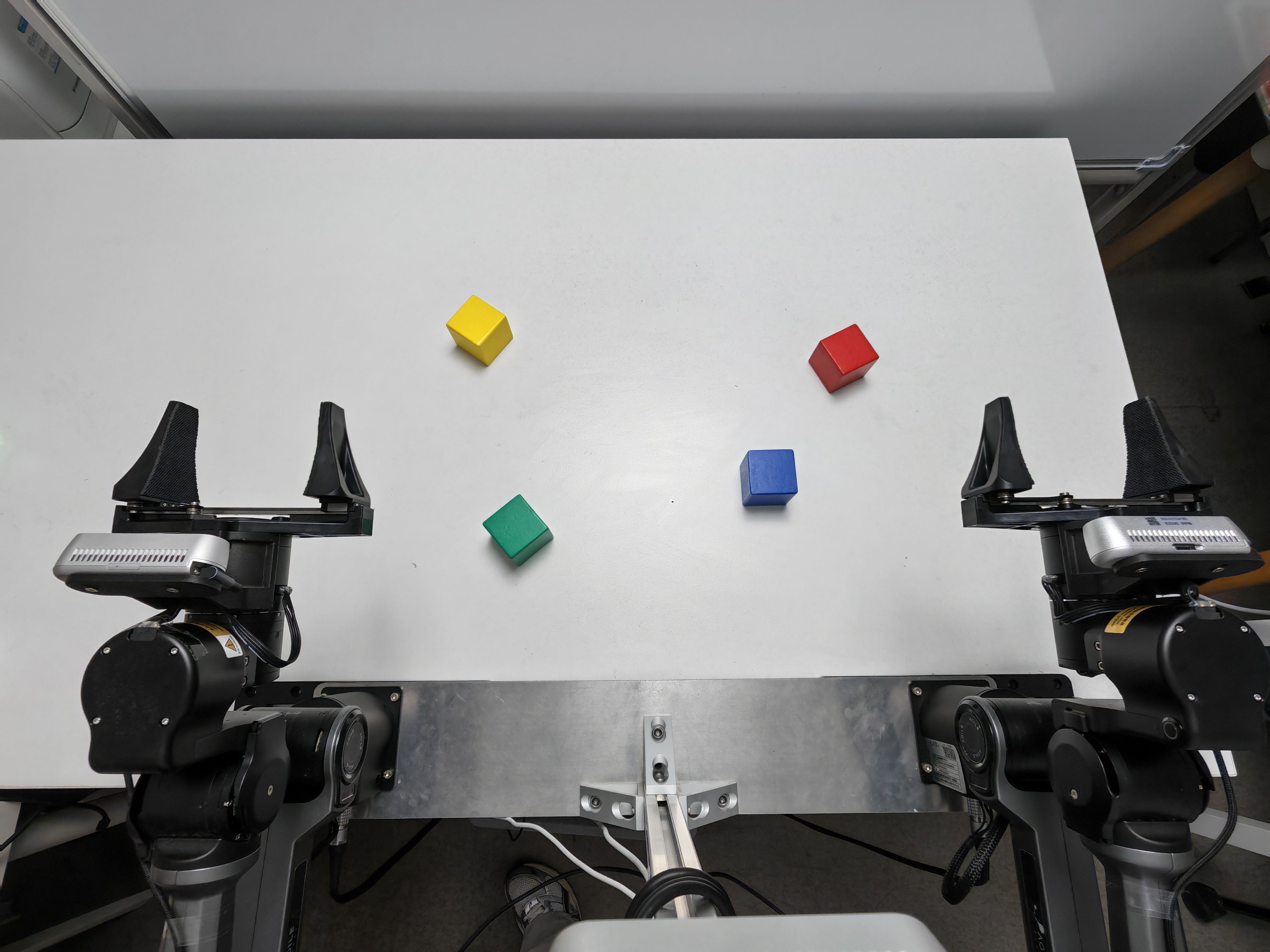}
    \end{tabular}
    \caption{Real-world scene overview for five tasks. Each row shows a task in the target-only condition (left) and in the distractor condition (right). The photographs show the task objects, placement targets where applicable, and robot grippers.}
    \label{fig:real-robot-task-scenes}
\end{figure}

The physical platform consists of two Agilex PiperX arms operating in a shared tabletop workspace. The scenes follow the task structure and object arrangements of the corresponding RoboTwin tasks. In the target condition, the workspace contains the objects and fixtures required for the task. The distractor condition adds irrelevant objects while preserving the task objective and the required objects. Both conditions are evaluated for every task and method in Table~\ref{tab:real-robot-main}. Figure~\ref{fig:real-robot-task-scenes} shows example layouts for the five tasks.

\subsection{Adaptation Data}
We fine-tune the real-world $\pi_{0.5}$ proposer on 500 real-robot demonstration trajectories. The Q evaluator is first trained on simulation MCTS data and subsequently adapted using the same 500 trajectories. Both methods use the same adapted policy checkpoint, while SVA additionally uses the adapted Q-model during evaluation.

\subsection{Evaluation and Aggregation}
For each of the five tasks, $\pi_{0.5}$ and $\pi_{0.5}$+SVA are evaluated under both scene conditions, with 25 trials for each task, condition, and method. Each entry in Table~\ref{tab:real-robot-main} reports the success rate together with the corresponding integer count.

\subsection{Deployment}
At each decision, the policy generates candidate continuous action chunks from the physical observation and task instruction. SVA scores the candidates with the adapted Q-model and executes a prefix of the selected chunk before observing again. Policy postprocessing maps commands to the robot's action representation; Q input normalization is used only for scoring. Both checkpoints remain fixed during the trials. The simulator supplies MCTS training supervision, while deployment uses physical observations and the learned models without online tree search.

\raggedbottom
\section{Method and Implementation Details}\label{app:sva_exp}

\paragraph{Problem formulation.}\label{app:problem_formulation}
We consider a language-conditioned MDP $\langle\mathcal S,\mathcal A,T,R,\gamma,\chi\rangle$, with state space $\mathcal S$, action space $\mathcal A$, transition function $T$, reward function $R$, discount factor $\gamma$, and language space $\chi$. The proposer and evaluator receive the observable representation of $s_t$ and instruction $l\in\chi$. Simulator states support branch restoration. One environment step denotes one atomic skill invocation or one application of a control vector. Instruction conditioning is omitted from some search expressions for readability.

\subsection{Q-Model Architecture}\label{app:q_architecture}
The evaluator uses \texttt{Qwen/Qwen3.5-0.8B} with hidden dimension 1024. For EmbodiedBench, the prompt contains the RGB observation, instruction, interaction history, and candidate skill names and arguments in execution order. A \texttt{<|VALUE|>} token follows the candidate and supplies its contextualized representation.

For continuous control, the visual and language prefix is followed by a proprioception token, temporal action tokens, and a learned Q readout embedding. Action dimensions are normalized using quantiles estimated from the training data and clipped to $[-1,1]$. Absolute joint space features concatenate each normalized target with its displacement from the normalized current joint state. Bridge end-effector actions use normalized targets alone. RoboTwin uses the fixed head, left-wrist, and right-wrist camera order, while SimplerEnv uses its primary camera.

The action encoder projects features to 256 dimensions, adds temporal position embeddings, and applies a two layer Transformer with four attention heads and feed forward dimension 1024. Outputs are projected to the backbone dimension, preserving one token per encoded action. An MLP encodes proprioception, and type embeddings distinguish the added inputs. The final readout is layer normalized and passed to five heads, each with
\[
\begin{gathered}
\texttt{Linear}(1024,512)\rightarrow\texttt{GELU}\rightarrow\texttt{Dropout}(0.1)\rightarrow\texttt{Linear}(512,1).
\end{gathered}
\]
The heads contain 2,626,565 parameters in total. LoRA, the heads, and the added continuous-input and readout modules are trained jointly. The backbone's base weights and the proposer remain frozen.

\subsection{Search Stage Data Collection}\label{app:search_details}
\paragraph{Selection and expansion.}
Search collects feedback through policy rollouts and credit assignment~\citep{lu2025onpolicydistillation}. Each edge stores $Q(s,a)$, prior $P(s,a)$, traversal count $n(s,a)$, and return-sample count $m(s,a)$. High-level reasoning uses PUCT. Continuous control first visits untried edges in descending order of prior, breaking ties randomly, and then uses UCT weighted by the prior:
\begin{equation}
\begin{aligned}
a_t&=\argmax_a\left[Q(s_t,a)+c_{\text{puct}}P(s_t,a)U(s_t,a)\right],\\
U(s_t,a)&=
\begin{cases}
\displaystyle\frac{\sqrt{n(s_t)}}{1+n(s_t,a)},&\text{high-level},\\[5pt]
\displaystyle\sqrt{\frac{\log(n(s_t)+1)}{n(s_t,a)}},&\text{continuous, }n(s_t,a)>0.
\end{cases}
\end{aligned}
\label{eq:puct_details}
\end{equation}
Here, $n(s_t)$ counts traversals through the parent and $c_{\text{puct}}$ controls exploration. Each simulation expands at most one node. Candidates become outgoing edges with priors from Eq.~\eqref{eq:candidate_prior}, and successors are obtained by executing the configured candidate prefix.

\paragraph{Rollout and backup.}
Simulator restoration or reset and replay enables repeated evaluation from a branch. The frozen proposer supplies rollout actions until termination or the configured limit, with zero continuation value at the cutoff. If an edge executes $\tau_t$ environment actions from $s_t$, including early termination, its backup is
\begin{equation}
G(s_t)=\sum_{j=0}^{\tau_t-1}\gamma^j r_{t+j}
+\gamma^{\tau_t}G(s_{t+\tau_t}).
\label{eq:chunk_backup}
\end{equation}
We increment $n(s,a)$ once per traversal. For each return sample reaching that edge, we update
\begin{equation}
m(s,a)\leftarrow m(s,a)+1,\qquad
Q(s,a)\leftarrow Q(s,a)+\frac{G(s)-Q(s,a)}{m(s,a)}.
\label{eq:backup_details}
\end{equation}
Separating the counts supports multiple rollout evaluations per traversal. Each ancestor receives its own discounted return, including path rewards. Targets reflect the evolving tree policy and frozen policy rollouts, with both successful and failed continuations retained.

\paragraph{Collection settings.}
Each sample stores the pre-action observation, instruction, candidate, empirical return, and search statistics, plus high-level interaction history or continuous control proprioception. Search runs on the training split. The SimplerEnv $\pi_0$ proposer uses \path{petkopetkov/INTACT-pi0-finetune-bridge}.

\begin{table}[H]
\centering
\caption{Search Stage settings. Cand. gives the sampling budget $M$, with $/N$ indicating the maximum retained medoids. Task-specific overrides are given below.}
\footnotesize
\setlength{\tabcolsep}{1.5pt}
\renewcommand{\arraystretch}{1.1}
\begin{tabular}{@{}lccccccc@{}}
\toprule
Benchmark & Proposer & Cand. & Sims. & Depth & $\gamma$ & $c_{\text{puct}}$ & \makecell{Training\\split} \\
\midrule
EB-Habitat & Qwen3.5 & 16 & 32 & 10 & 0.98 & 2.0 & Ep.~1--30 \\
EB-Navigation & Qwen3.5 & 16 & 32 & 10 & 0.98 & 2.0 & Ep.~1--36 \\
SimplerEnv & $\pi_0$ & 32 & 256 & 15 & 0.99 & 4.0 & 15 seeds \\
RoboTwin 2.0 & $\pi_{0.5}$ & 32/8 & 128 & 8 & 0.99 & 3.0 & \makecell{30 episodes\\per task} \\
\bottomrule
\end{tabular}
\label{tab:app_search}
\end{table}

RoboTwin uses one root search per episode and executes up to 24 control steps per search or rollout decision. Search and rollout depths are 8 by default, 10 for \texttt{place\_bread\_skillet} and \texttt{scan\_object}, and 15 for \texttt{stack\_blocks\_two}, with step caps 192, 240, and 360. The last task uses 256 simulations and $\gamma=0.995$. Each endpoint receives one rollout initially and one on revisits, sampling one candidate per rollout decision. Training episodes are selected by scanning seeds from 100000.

\subsection{Value Stage Training}\label{app:value_training}
We train separate evaluators for EB-Habitat, EB-Navigation, SimplerEnv, and RoboTwin. High-level return targets are standardized using training-set statistics. continuous control targets use raw collected returns. Each head receives its own Smooth-$L1$ loss, averaged over heads and samples:
\begin{equation}
\begin{aligned}
\mathcal L_{\mathrm{reg}}(\phi)&=\frac1B\sum_{i=1}^{B}
\begin{cases}
\tfrac12(\epsilon^{(i)})^2,&|\epsilon^{(i)}|<1,\\
|\epsilon^{(i)}|-\tfrac12,&\text{otherwise},
\end{cases}\\
\epsilon^{(i)}&=\hat Q_\phi^{(i)}(s,a;l)-Q(s,a;l).
\end{aligned}
\end{equation}
Here, $Q(s,a;l)$ uses the target scaling specified above. The continuous trainer groups sibling candidates from the same search node and optionally adds a pairwise loss:
\begin{equation}
\mathcal L=w_{\mathrm{reg}}\mathcal L_{\mathrm{reg}}
+w_{\mathrm{pair}}\mathcal L_{\mathrm{pair}}.
\label{eq:combined_q_loss}
\end{equation}
For each sibling pair, sigmoid-transformed prediction and target differences use the same temperature $T_{\mathrm{pair}}$. Binary cross-entropy is averaged over heads, non-tied pairs, and groups with informative pairs. Pairs within the target-gap threshold are omitted, with zero pairwise loss when none remain. Setting $w_{\mathrm{pair}}=0$ recovers regression. Heads share targets without head-specific bootstrap masks. All targets remain fixed, without temporal-difference bootstrapping.

\paragraph{Optimization settings.}
All evaluators use AdamW for 5 epochs, bfloat16 precision, and gradient clipping 1.0. EmbodiedBench and SimplerEnv use batch size 8, learning rate $10^{-4}$, and LoRA rank 16 with $\alpha=32$. LoRA targets \texttt{q\_proj}, \texttt{k\_proj}, \texttt{v\_proj}, \texttt{o\_proj}, \texttt{gate\_proj}, \texttt{up\_proj}, and \texttt{down\_proj}.

RoboTwin trains one evaluator jointly on all 12 tasks, using
14-dimensional end-effector features. It uses learning rate $5\times10^{-5}$, weight decay 0.01, 32 sibling groups per device, one accumulation step, and seed 0. The loss settings are $(w_{\mathrm{reg}},w_{\mathrm{pair}},T_{\mathrm{pair}})=(1,0.02,0.2)$ and tie threshold 0.02. LoRA uses rank 8, $\alpha=16$, and targets \texttt{q\_proj}, \texttt{k\_proj}, \texttt{v\_proj}, \texttt{o\_proj}, \texttt{in\_proj\_qkv}, \texttt{in\_proj\_z}, and \texttt{out\_proj}. Both settings use LoRA dropout 0.05 without bias training. RoboTwin shares observation prefixes during training and disables gradient checkpointing.

\subsection{Act Stage Evaluation}\label{app:act_details}
\paragraph{Candidates and priors.}
We generate $M$ samples and retain $N\leq M$ candidates. High-level candidates merge identical ordered skill sequences, including arguments.Continuous candidates are either scored directly with uniform priors $p=1/N$ or clustered according to the distance between their postprocessed execution prefixes. For chunks in joint space, the clustering features concatenate displacements scaled by fixed action scales and their temporal increments, with configurable weights for gripper motion and increments. Average linkage clustering retains sampled medoids. Exact duplicates are merged, so the retained count can be below the requested budget.

Let $\mathcal C_i$ contain the samples assigned to $a^{(i)}$, by exact equality for skills or clustering for continuous chunks. The priors are
\begin{equation}
p(a^{(i)}\mid s_t;l)=
\begin{cases}
\displaystyle\frac{|\mathcal C_i|}{M},&\text{high-level},\\[6pt]
\displaystyle(1-\varepsilon)
\frac{(|\mathcal C_i|+\alpha)^\beta}
{\sum_{j=1}^{N}(|\mathcal C_j|+\alpha)^\beta}
+\frac{\varepsilon}{N},&\text{clustered continuous},
\end{cases}
\label{eq:candidate_prior}
\end{equation}
where $\alpha\geq0$, $0<\beta\leq1$, and $0\leq\varepsilon\leq1$. Continuous priors estimate cluster mass. Search uses the same construction. For fixed $N$, $\log(Np)$ gives the same selection as $\log p$ in Eq.~\eqref{eq:act_selection} and vanishes for uniform priors.

\paragraph{Scoring and execution.}
The evaluator uses an ensemble of $H=5$ Q-heads. Let
$\hat Q_\phi^{(i)}(s,a;l)$ denote the prediction of the $i$-th head.
We compute the ensemble mean and disagreement as
\begin{equation}
\begin{aligned}
\mu_\phi(s,a;l)
&= \frac{1}{H}\sum_{i=1}^{H}
\hat Q_\phi^{(i)}(s,a;l),\\
\sigma_\phi(s,a;l)
&= \sqrt{
\frac{1}{H}\sum_{i=1}^{H}
\left(
\hat Q_\phi^{(i)}(s,a;l)
-
\mu_\phi(s,a;l)
\right)^2
}.
\end{aligned}
\label{eq:ensemble_statistics}
\end{equation}
Prediction length, Q horizon, and executed prefix length are distinct. The checkpoint fixes the Q horizon, action dimensions, camera order, and normalization. Q scoring pads or trims chunks, while execution uses the original postprocessed actions. Relative policy outputs are postprocessed once using the proposal-generation observation as the anchor. The agent executes the selected prefix and replans, stopping on success, invalid feedback, or the episode limit. In Table~\ref{tab:app_act}, $\lambda_u=\lambda_1$ and $\lambda_p=\lambda_2$.

\begin{table}[H]
\centering
\caption{Configured Act Stage settings. Candidates gives the sampling budget $M$ and, for clustered candidates, the maximum retained count $N$. Duplicate removal can reduce $N$. Horizon is the executed prefix length. EmbodiedBench uses temperature 0.7.}
\small
\setlength{\tabcolsep}{5pt}
\renewcommand{\arraystretch}{1.1}
\begin{tabular}{lccccc}
\toprule
Benchmark & Cand. & Horizon & $\lambda_u$ & $\lambda_p$ & Evaluation split \\
\midrule
EB-Habitat & 32 & 10 & 0.1 & 0.1 & Episodes 31--50 \\
EB-Navigation & 32 & 6 & 0.1 & 0.1 & Episodes 37--60 \\
SimplerEnv & 16 & 4 & 0.1 & 0 & 9 held-out seeds \\
RoboTwin 2.0 & 32/16 & 50 & 0.2 & 0.001 & \makecell{50 episodes per task} \\
\bottomrule
\end{tabular}
\label{tab:app_act}
\end{table}

\subsubsection{Efficient Inference}\label{app:pi05-act-latency}
\paragraph{Shared Prefix Candidate Generation.}
For a single observation and instruction, $\pi_{0.5}$ encodes the visual--language prefix once and reuses its key--value cache across candidate trajectories. Independent noise samples initialize the candidates, which undergo batched action denoising. Each denoising iteration still computes attention over the action suffix and feedforward operations for every candidate. Prefix reuse therefore amortizes the fixed encoding cost, while the computation required for the candidates grows with $M$. The generated chunks pass through policy postprocessing before Q evaluation and selection.

\paragraph{Q Scoring and CUDA Graph Replay.}
The Q-model shares the observation prefix, including visual, language, and proprioceptive inputs, within each request and processes candidate action suffixes in a batch. In the measured implementation, CUDA Graph captures the computation of the shared decoder for a fixed input shape. Subsequent requests copy current inputs into the capture buffers and replay the graph, reducing scheduling and short kernel launch overhead. Replay requires the candidate count and token dimensions to match the captured shape. Observation features are refreshed for each request. The graph reuses an execution schedule, while the decoder evaluates the current observation and candidate actions. Q head predictions then determine the selected chunk for execution.

\begin{figure}[tb]
    \centering
    \includegraphics[width=\linewidth]{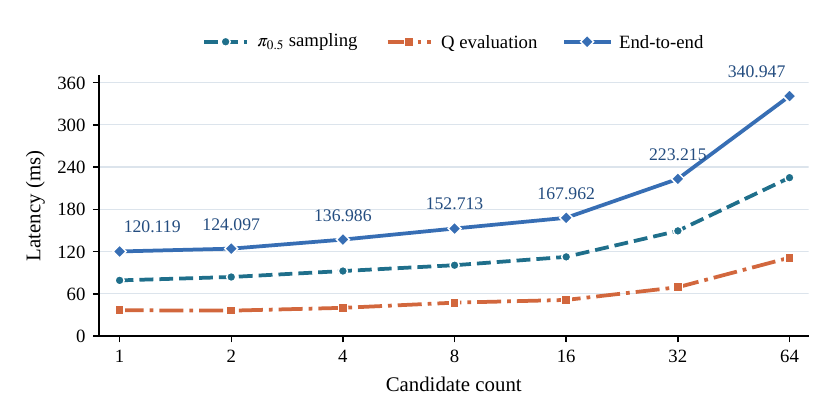}
    \caption{Mean latency of the accelerated $\pi_{0.5}$ Act pipeline on a single NVIDIA GeForce RTX 4090 GPU. All sampled candidates are scored ($M=N$). Curves show $\pi_{0.5}$ sampling, Q evaluation with CUDA Graph enabled, and the end-to-end measurement. The candidate axis uses a logarithmic scale with base 2.}
    \label{fig:pi05-act-latency}
\end{figure}

\paragraph{Mean Latency Across Candidate Budgets.}
We measure latency on a single NVIDIA GeForce RTX 4090 GPU. We vary the sampled and scored candidate counts together, $M=N\in\{1,2,4,8,16,32,64\}$, retaining all candidates. Figure~\ref{fig:pi05-act-latency} reports mean latency for $\pi_{0.5}$ sampling, Q evaluation with CUDA Graph enabled, and the end-to-end Act measurement. From 1 to 64 candidates, sampling increases from $79.027$ to $224.994$\,ms, and Q evaluation increases from $36.578$ to $111.260$\,ms. End-to-end latency rises from $120.119$ to $340.947$\,ms, a $2.84\times$ increase over the tested $64\times$ candidate range.

%================================================
%================================================
\section{Baselines}\label{app:baselines}
\subsection{RoboTwin Comparison}
For the main RoboTwin comparison, both $\pi_{0.5}$ and FastWAM use checkpoints from the official RoboTwin 2.0 \texttt{cotrain\_ckpt} release, under \texttt{Pi\_05} and \texttt{FastWAM}, respectively.\footnote{\url{https://huggingface.co/datasets/TianxingChen/RoboTwin2.0/tree/main/cotrain_ckpt}} The direct policy baselines execute chunks from these checkpoints. Their SVA variants use the corresponding frozen policy for proposals and a learned evaluator for selection. FastWAM retains video co-training during learning and uses direct action generation at inference~\citep{yuan2026fastwam}; SVA treats its action chunks through the same continuous candidate interface.

The additional $\pi_{0.5}$ baselines use two distinct selection signals. V-GPS learns an action-value function through offline RL and uses it to select among actions sampled from the policy~\citep{nakamoto2025steering}. TACO learns a pseudo-count estimator from policy representations and selects the candidate with the largest estimated count~\citep{yang2025steering}. Its learned pseudo-count is different from SVA's empirical candidate prior. The method names in Table~\ref{tab:robotwin-main} denote the corresponding combinations of selectors and proposers.

\subsection{SimplerEnv Comparison}
For SimplerEnv, we use the released RoboMonkey verifier, \path{robomonkey-vla/monkey-verifier-7b}, as a frozen verifier. The policy samples multiple 7-DoF WidowX action chunks. For each chunk, we query the verifier for every action and use the mean reward across the chunk as its score. For scoring, 5 action proposals are generated by $\pi_0$, and the remaining proposals are sampled from a Gaussian distribution fitted to the policy outputs. The grasp state is determined separately by majority voting. The candidate with the highest mean score is selected for execution.

%========================================================
% Keep appendix float pages top-aligned without stretching table gaps.
\makeatletter
\setlength{\@fptop}{0pt}
\setlength{\@fpsep}{14pt}
\setlength{\@dblfptop}{0pt}
\setlength{\@dblfpsep}{14pt}
\makeatother
\section{Detailed Pass@k Results}\label{app:pass_k}
Table~\ref{tab:passk_task_level} reports task-level Pass@$k$ results on representative manipulation benchmarks. The gap between Pass@1 and larger $k$ shows that repeated full rollouts can recover successes missed by a single attempt. It motivates evaluating candidate actions, whose benefit is measured separately in the SVA experiments.

\begin{table}[H]
\centering
\caption{Task-level Pass@$k$ results across manipulation benchmarks using popular VLA models.}
\label{tab:passk_task_level}
\resizebox{\linewidth}{!}{
\begin{tabular}{llcccccc}
\toprule
\textbf{Benchmark} & \textbf{Task} & \textbf{Pass@1} & \textbf{Pass@2} & \textbf{Pass@4} & \textbf{Pass@8} & \textbf{Pass@16} & \textbf{Pass@32} \\
\midrule
\multirow{4}{*}{\makecell[l]{Libero Long\\(OpenVLA)}}
& Pick up Book             & 0.8440 & 0.9674 & 0.9981 & 1.0000 & 1.0000 & 1.0000 \\
& Soup and Sauce in Basket & 0.6460 & 0.8467 & 0.9614 & 0.9970 & 1.0000 & 1.0000 \\
& Put Mug on Plate         & 0.4420 & 0.6600 & 0.8390 & 0.9383 & 0.9856 & 0.9996 \\
& Moka Pots on Stove       & 0.2640 & 0.4459 & 0.6682 & 0.8649 & 0.9763 & 0.9998 \\
\midrule
\multirow{4}{*}{\makecell[l]{Simpler\\(OpenVLA)}}
& Spoon on Towel           & 0.1700 & 0.2571 & 0.3646 & 0.4949 & 0.6469 & 0.8087 \\
& Carrot on Plate          & 0.0740 & 0.1385 & 0.2453 & 0.3993 & 0.5936 & 0.8284 \\
& Stack Cubes              & 0.0260 & 0.0507 & 0.0966 & 0.1767 & 0.3057 & 0.5074 \\
& Eggplant in Basket       & 0.1080 & 0.2016 & 0.3541 & 0.5639 & 0.7873 & 0.9568 \\
\midrule
\multirow{4}{*}{\makecell[l]{RoboTwin\\($\pi_{0.5}$)}}
& Turn Switch              & 0.3100 & 0.4950 & 0.6851 & 0.8248 & 0.9205 & 0.9858 \\
& Move Stapler Pad         & 0.1460 & 0.2651 & 0.4455 & 0.6665 & 0.8638 & 0.9787 \\
& Click Bell               & 0.6600 & 0.8078 & 0.8927 & 0.9377 & 0.9694 & 0.9958 \\
& Rotate QRCode            & 0.3560 & 0.5482 & 0.7346 & 0.8747 & 0.9588 & 0.9957 \\
\bottomrule
\end{tabular}
}
\end{table}

\section{More Results on Case Study}\label{app:case_study}

We provide additional qualitative results for two representative cases, each comparing the base policy with a successful SVA rollout.

\begin{figure}[p]
    \centering
    \setlength{\tabcolsep}{3pt}
    \renewcommand{\arraystretch}{1.0}
    \setlength{\fboxsep}{1pt}\setlength{\fboxrule}{0.6pt}

    \begin{tabularx}{\textwidth}{YYYY}
        \toprule
        \multicolumn{4}{>{\columncolor{white}}c}{
            \makecell[c]{
                \large\textbf{CASE 1}\\
                \emph{On the sofa there's an apple, but instead find a plate and move it to the brown table.}
            }
        } \\
        \midrule
        \multicolumn{4}{>{\columncolor{lightRed}}c}{
            \textcolor{baseRed}{\textbf{Base Policy} \quad \ding{55}~Failure}
        } \\

        \fcolorbox{baseRed}{white}{\includegraphics[width=0.91\linewidth]{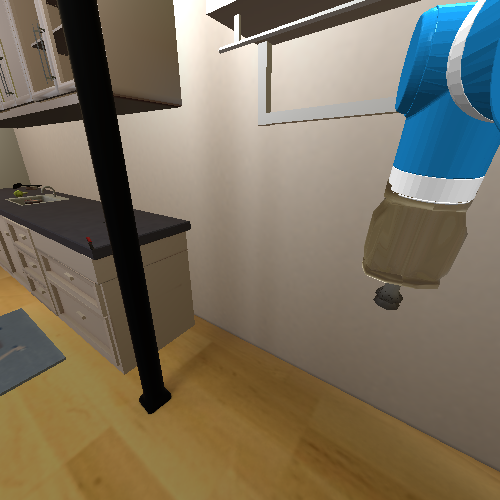}}\newline
        {\footnotesize Step 0: navigate to sofa}
        &
        \fcolorbox{baseRed}{white}{\includegraphics[width=0.91\linewidth]{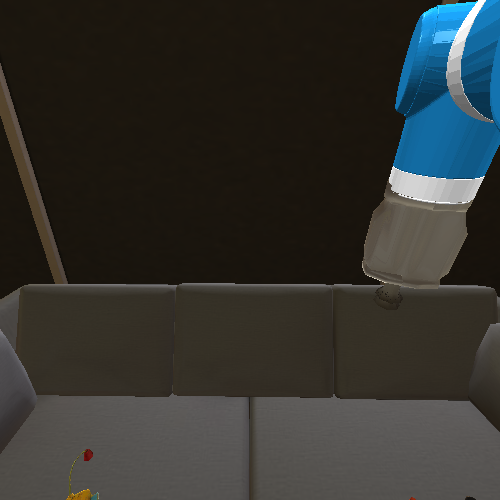}}\newline
        {\footnotesize Step 1: follow distractor}
        &
        \fcolorbox{baseRed}{white}{\includegraphics[width=0.91\linewidth]{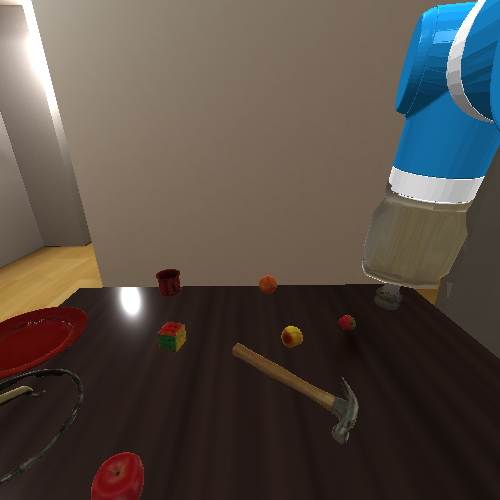}}\newline
        {\footnotesize Step 6: invalid recovery}
        &
        \fcolorbox{baseRed}{white}{\includegraphics[width=0.91\linewidth]{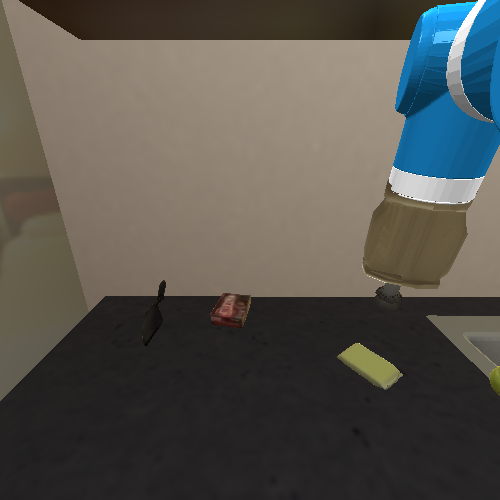}}\newline
        {\footnotesize Step 25: failure}
        \\[0.3em]

        \multicolumn{4}{>{\columncolor{lightBlue}}c}{
            \textcolor{svaBlue}{\textbf{SVA (Ours)} \quad \ding{51}~Success}
        } \\

        \fcolorbox{svaBlue}{white}{\includegraphics[width=0.91\linewidth]{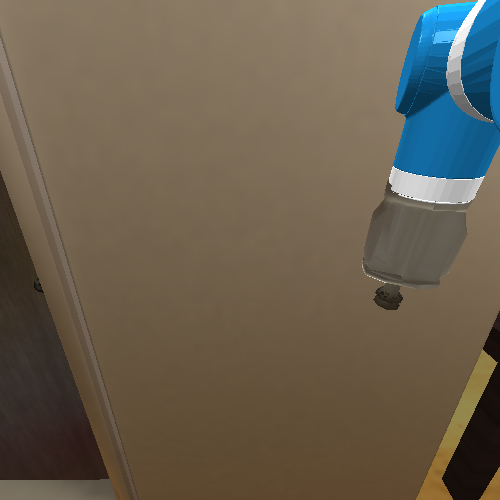}}\newline
        {\footnotesize Step 0: nav. to kitchen}
        &
        \fcolorbox{svaBlue}{white}{\includegraphics[width=0.91\linewidth]{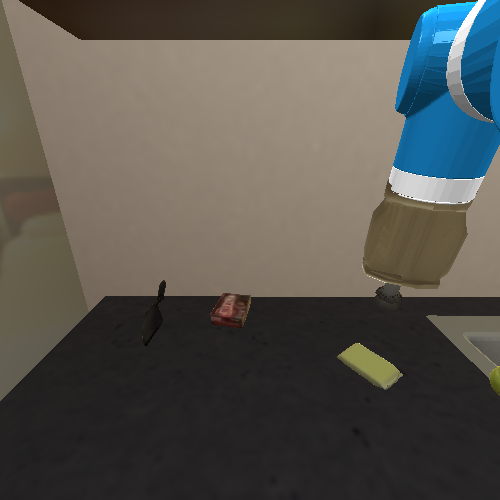}}\newline
        {\footnotesize Step 1: find plate}
        &
        \fcolorbox{svaBlue}{white}{\includegraphics[width=0.91\linewidth]{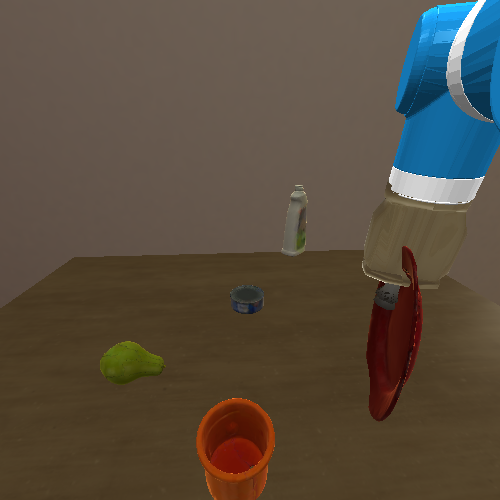}}\newline
        {\footnotesize Step 3: move to target}
        &
        \fcolorbox{svaBlue}{white}{\includegraphics[width=0.91\linewidth]{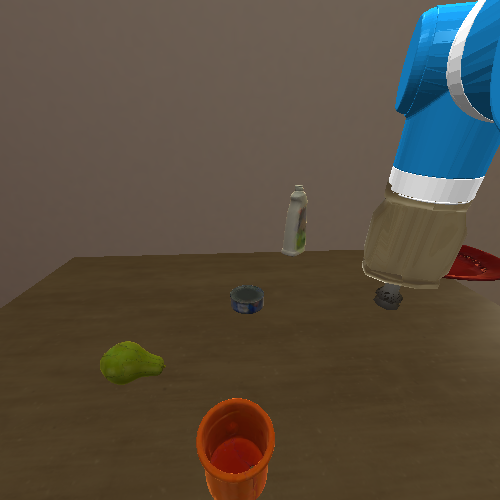}}\newline
        {\footnotesize Step 4: success}
        \\
        \bottomrule
    \end{tabularx}

    \caption{
    \textbf{Case 1 -- distractor-aware instruction following.} The base policy follows the salient but irrelevant apple-on-sofa clause and fails after repeated invalid recovery actions. SVA selects a plate-centric plan and completes the task in 4 steps.
    }
    \label{fig:case1-study}
% Keep both case studies on one figure page; captions retain separate numbers.
    \par\vspace{6pt}
    \centering
    \setlength{\tabcolsep}{3pt}
    \renewcommand{\arraystretch}{1.0}
    \setlength{\fboxsep}{1pt}\setlength{\fboxrule}{0.6pt}

    \begin{tabularx}{\textwidth}{YYYY}
        \toprule
        \multicolumn{4}{>{\columncolor{white}}c}{
            \makecell[c]{
                \large\textbf{CASE 2}\\
                \emph{Find a wrench and move it to the right of the sink.}
            }
        } \\
        \midrule
        \multicolumn{4}{>{\columncolor{lightRed}}c}{
            \textcolor{baseRed}{\textbf{Base Policy} \quad \ding{55}~Failure}
        } \\

        \fcolorbox{baseRed}{white}{\includegraphics[width=0.91\linewidth]{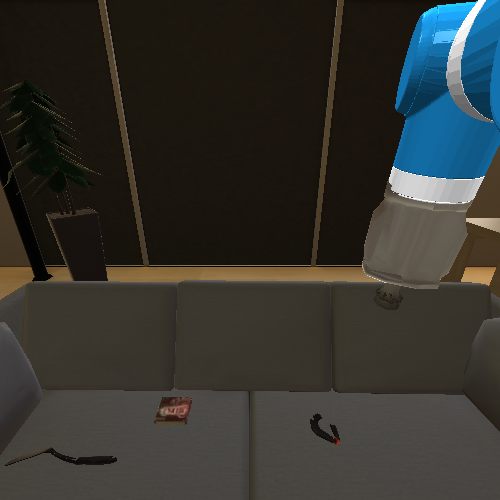}}\newline
        {\footnotesize Step 1: wrong search}
        &
        \fcolorbox{baseRed}{white}{\includegraphics[width=0.91\linewidth]{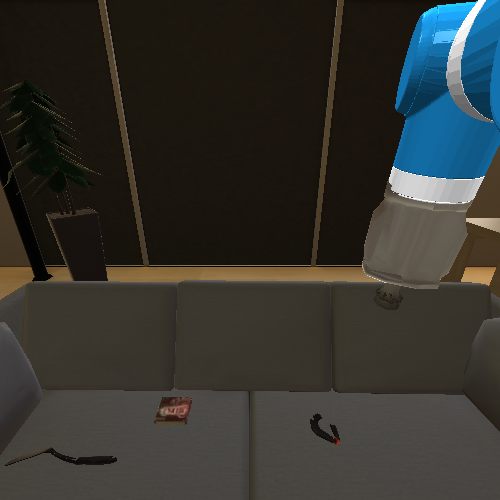}}\newline
        {\footnotesize Step 2: invalid pickup}
        &
        \fcolorbox{baseRed}{white}{\includegraphics[width=0.91\linewidth]{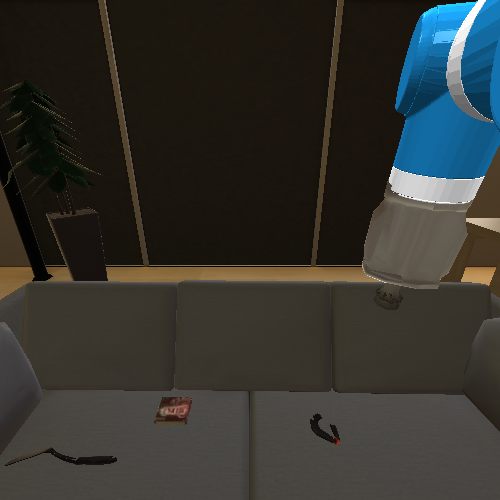}}\newline
        {\footnotesize Step 10: repeated failure}
        &
        \fcolorbox{baseRed}{white}{\includegraphics[width=0.91\linewidth]{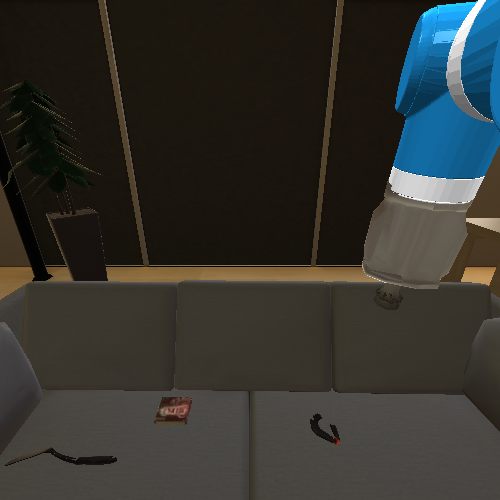}}\newline
        {\footnotesize Step 20: failure}
        \\[0.3em]

        \multicolumn{4}{>{\columncolor{lightBlue}}c}{
            \textcolor{svaBlue}{\textbf{SVA (Ours)} \quad \ding{51}~Success}
        } \\

        \fcolorbox{svaBlue}{white}{\includegraphics[width=0.91\linewidth]{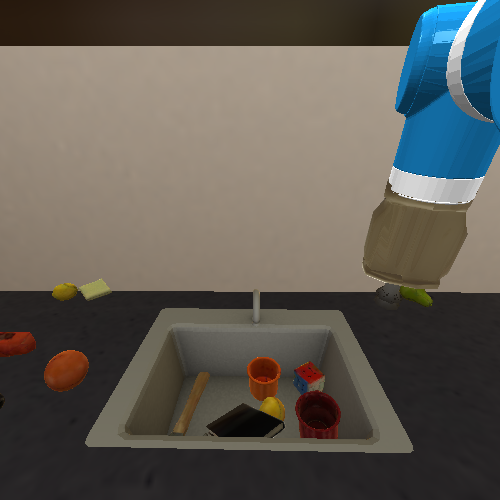}}\newline
        {\footnotesize Step 1: nav. to sink}
        &
        \fcolorbox{svaBlue}{white}{\includegraphics[width=0.91\linewidth]{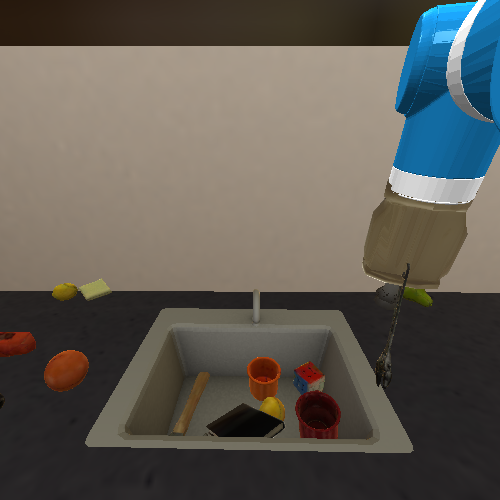}}\newline
        {\footnotesize Step 2: pick wrench}
        &
        \fcolorbox{svaBlue}{white}{\includegraphics[width=0.91\linewidth]{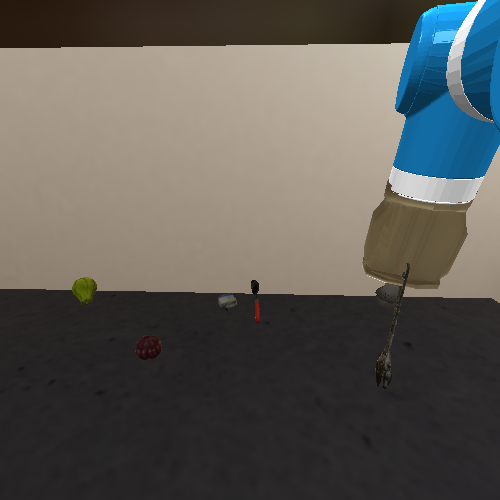}}\newline
        {\footnotesize Step 3: move right}
        &
        \fcolorbox{svaBlue}{white}{\includegraphics[width=0.91\linewidth]{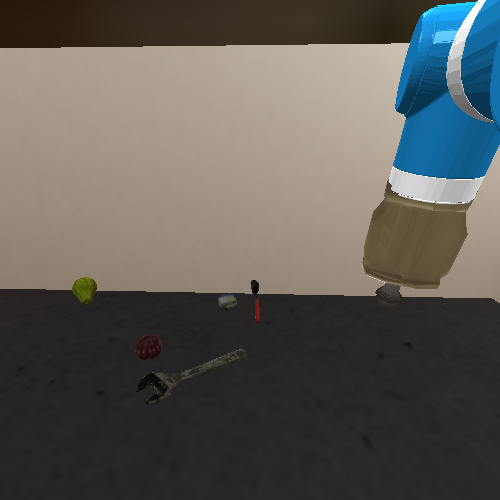}}\newline
        {\footnotesize Step 4: success}
        \\
        \bottomrule
    \end{tabularx}

    \caption{
    \textbf{Case 2 -- spatial-relation grounding.} For the instruction \emph{Find a wrench and move it to the right of the sink}, the base policy searches an incorrect location and issues repeated invalid pickup actions, while SVA grounds the spatial relation and completes the rearrangement.
    }
    \label{fig:case2-study}
\end{figure}

\paragraph{Case 1: distractor-aware instruction following.}
The instruction is: \emph{On the sofa there's an apple, but instead find a plate and move it to the brown table.}
As shown in Fig.~\ref{fig:case1-study}, the base policy is distracted by the sofa clause and fails after repeated invalid actions, while SVA follows the plate-centric plan and completes the task.

\paragraph{Case 2: spatial-relation grounding.}
We further examine the instruction: \emph{Find a wrench and move it to the right of the sink.}
As shown in Fig.~\ref{fig:case2-study}, the base policy repeatedly navigates to the sofa and attempts to pick up the wrench, producing no task progress. In contrast, SVA selects the spatially correct sequence: navigate to the sink, pick up the wrench, move to the right counter, and place it there.

\clearpage
\flushbottom
\makeatletter
\setlength{\@fptop}{0pt plus 1fil}
\setlength{\@fpsep}{0.75\baselineskip plus 2fil}
\setlength{\@dblfptop}{0pt plus 1fil}
\setlength{\@dblfpsep}{0.75\baselineskip plus 2fil}
\makeatother
\end{document}